\documentclass{article}

\usepackage[final,nonatbib]{neurips_2022}

\usepackage[square,numbers]{natbib}
\usepackage{graphicx}
\usepackage{subfigure}
\usepackage[utf8]{inputenc} 
\usepackage[T1]{fontenc}    
\usepackage[OT1]{fontenc} 
\usepackage{url}            
\usepackage{booktabs}       
\usepackage{amsfonts}       
\usepackage{nicefrac}       
\usepackage{microtype}      
\usepackage{xcolor}         

\usepackage{wrapfig}
\usepackage{graphicx}
\usepackage{subfigure}
\usepackage{amsmath}
\usepackage{amssymb}
\usepackage{mathtools}
\usepackage{amsthm}
\usepackage{xcolor}
\usepackage{multirow}
\usepackage{makecell}

\usepackage{colortbl,color}
\definecolor{greyC}{RGB}{180,180,180}
\definecolor{greyL}{RGB}{235,235,235}

\usepackage{algorithm}
\usepackage[noend]{algpseudocode}

\definecolor{mydarkblue}{rgb}{0,0.08,0.45}
\usepackage[colorlinks, citecolor=mydarkblue]{hyperref}

\title{Watermarking for Out-of-distribution Detection}

\author{%
Qizhou Wang$^{1}$\thanks{Equal contributions.} \quad Feng Liu$^{2}$\footnotemark[1] \quad Yonggang Zhang$^{1}$ \quad Jing Zhang$^{3}$ \\ \textbf{Chen Gong$^{4,5}$ \quad Tongliang Liu$^{6}\thanks{Correspondence to Bo Han (bhanml@comp.hkbu.edu.hk) and Tongliang Liu (tongliang.liu@sydney.edu.au).}$ \quad Bo Han$^{1\dag}$} \\
  $^1$Department of Computer Science, Hong Kong Baptist University \\
  $^2$School of Mathematics and Statistics, The University of Melbourne \\
  $^3$School of Computer Science, The University of Sydney \\
  $^4$PCA Lab, Key Lab of Intelligent Perception and Systems for High-Dimensional Information of MoE \\
  $^5$Jiangsu Key Lab of Image and Video Understanding for Social Security, \\
  School of Computer Science and Engineering, Nanjing University of Science and Technology\\
  $^6$TML Lab, The University of Sydney \\
  \textnormal{\{csqzwang, csygzhang, bhanml\}@comp.hkbu.edu.hk} \\ \textnormal{fengliu.ml@gmail.com} \quad
  \textnormal{chen.gong@njust.edu.cn}\\
  \textnormal{\{jing.zhang1, tongliang.liu\}@sydney.edu.au}
}

\begin{document}

\maketitle

\begin{abstract}
  \emph{Out-of-distribution} (OOD) detection aims to identify OOD data based on representations extracted from well-trained deep models. However, existing methods largely ignore the \emph{reprogramming} property of deep models and thus may not fully unleash their intrinsic strength: \emph{without} modifying parameters of a well-trained deep model, we can reprogram this model for a new purpose via data-level manipulation (e.g., adding a specific feature perturbation to the data). This property motivates us to reprogram a classification model to excel at OOD detection (a new task), and thus we propose a general methodology named \emph{watermarking} in this paper. Specifically, we learn a unified pattern that is superimposed onto features of original data, and the model's detection capability is largely boosted after watermarking. Extensive experiments verify the effectiveness of watermarking, demonstrating the significance of the reprogramming property of deep models in OOD detection. The code is publicly available at: \href{https://github.com/QizhouWang/watermarking}{{github.com/qizhouwang/watermarking}}.
\end{abstract}

\section{Introduction}

Deep learning systems in an open world often encounter \emph{out-of-distribution} (OOD) inputs whose label spaces are disjoint with that of training data, known as \emph{in-distribution} (ID) data. For safety-critical applications, deep models should make reliable predictions for ID data, meanwhile detecting OOD data and avoiding making predictions for the detected ones. This leads to the OOD detection task~\cite{lee2018simple,nalisnickdo19,RenLFSPDDL19,vernekar2019out}, which has attracted intensive attention in the real world.

Identifying OOD data remains non-trivial since deep models can be overconfident with them~\cite{nguyen2015deep}. As a promising technique, the \emph{classification-based} OOD detection~\cite{yang2021generalized} relies on various scoring functions derived by classification models well trained with ID data (i.e., well-trained models), taking those inputs with small scores as OOD cases. In general, the scoring functions can be defined by logit outputs~\cite{hendrycks2016baseline,liu2020energy}, gradients~\cite{huang2021importance}, and embedding features~\cite{lee2018simple,sastry2019detecting}. Without interfering with the well-trained models or requiring extra computation, they exploit the inherent capability of models learned from only ID data. In general, these advantages can be critical in reality, where the cost of re-training is prohibitively high and the acquisition of true OOD data is very difficult~\cite{yang2021generalized}. 




Although promising progress has been achieved, previous works largely ignore the \emph{reprogramming} property~\cite{elsayed2018adversarial} of deep models: a well-trained model can be repurposed for a new task by a proper transformation of original inputs (e.g., a universal feature perturbation), without modifying any model parameter. For example, a model pre-trained on ImageNet~\cite{deng2009imagenet} dataset can be reprogrammed for classifying biomedical images~\cite{TsaiCH20}.
This property indicates the possibility of making a well-trained model adapt for effective OOD detection, motivating us to make the \emph{first} attempt to investigate if the reprogramming property of deep models can help to address OOD detection, \emph{i.e., can we reprogram well-trained deep models for OOD detection (a new task)?}

\begin{wrapfigure}{r}{0.5\textwidth}
    \centering
    \includegraphics[width=\linewidth]{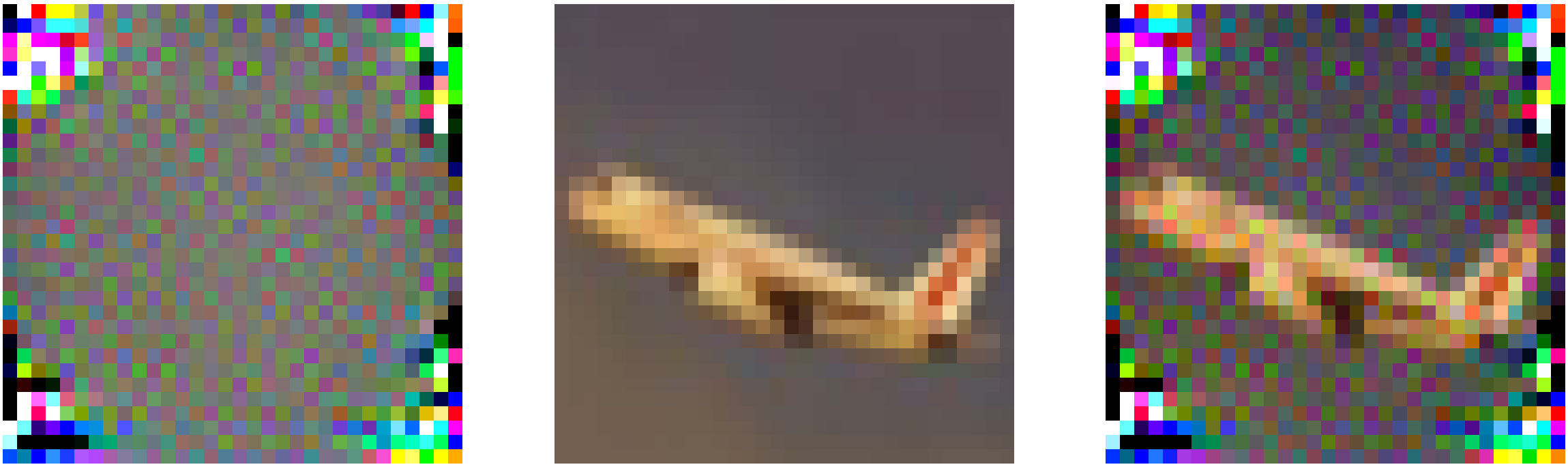}
    \vspace{1.5pt}
    \caption{Watermarking on CIFAR-$10$~\cite{krizhevsky2009learning} with free energy scoring~\cite{liu2020energy}. The left figure is the learned watermark; the middle figure is an original input; the right figure is the watermarked result.} 
    \label{fig: wp_example}
\end{wrapfigure}
\begin{figure*}[t]
    \centering
    \subfigure[before watermarking]{
    \centering  
    \begin{minipage}[t]{0.45\textwidth}
        \centering
	   \includegraphics[width=\linewidth]{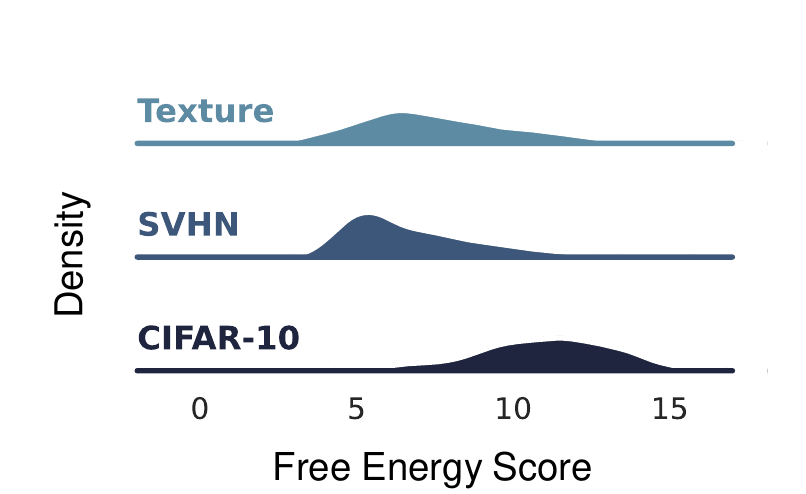}
        \centering  \label{fig: density_wo}
    \end{minipage}} 
    ~~~~~
    \subfigure[after watermarking]{
    \centering  
    \begin{minipage}[t]{0.45\textwidth}
        \centering  
        \includegraphics[width=\linewidth]{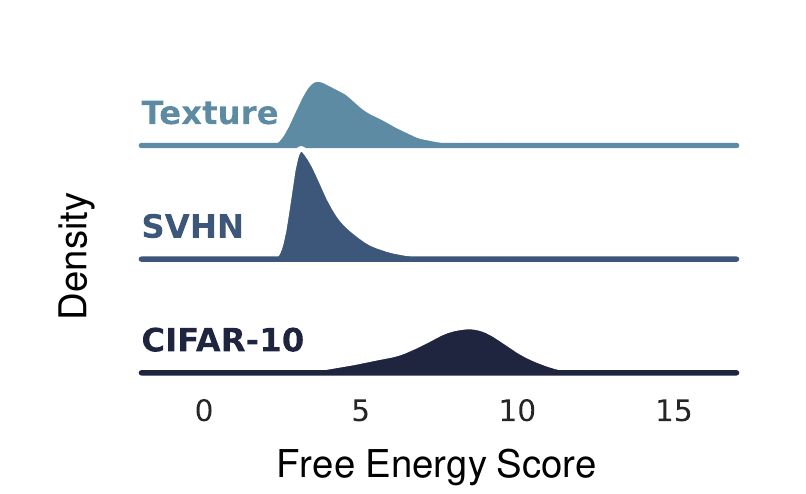}
        \centering \label{fig: density_w}
    \end{minipage}}
    \caption{
    Experimental results before (a) /after (b) watermarking with CIFAR-$10$ being the ID dataset, SVHN and Texture being the OOD datasets. Data with large (small) OOD scores should be taken as ID (OOD) data, and a larger distribution gap of scoring between ID and OOD data ensures a better detection performance. After watermarking, the gap between ID and OOD data is enlarged, demonstrating the improved capability of the original model in OOD detection. The horizontal axes are ignored for illustration, please refer to Figure~\ref{fig: kde2} for a completed version.
    }
    \label{fig:kde}
\end{figure*}

In this paper, we propose a novel method, \textbf{watermarking}, to \emph{reprogram} a well-trained model by adding a watermark to original inputs, making the model can help detect OOD data well. The \emph{watermark} has the same shape with original inputs, which is a static pattern that can be added for test-time inputs (cf., Figure~\ref{fig: wp_example}). The pre-defined scoring strategy (e.g., the free energy scoring \cite{liu2020energy}) is expected to be enhanced, with an enlarged gap of OOD scores between the watermarked ID and OOD data (cf., Figure~\ref{fig:kde}). 


It is non-trivial to find the proper watermark due to our lack of knowledge about unseen OOD data in advance. To address the issue, we propose a learning framework for effective watermarking. The insight is to make a well-trained model produce high scores for watermarked ID inputs meanwhile regularize the watermark such that the model will return low confidence without perceiving ID pattern. In this case, the model will have a relatively high score for a watermarked ID input, while the score remains low for OOD data (cf., Figure~\ref{fig:kde}). The reason is that the model encounters a watermarked input but not seeing any ID pattern. In our realization, we adopt several representative scoring strategies, devising specified learning objectives and proposing a reliable optimization algorithm to learn an effective watermark. 



To understand our watermarking, Figure~\ref{fig: wp_example} depicts the watermark learned on CIFAR-$10$~\cite{krizhevsky2009learning} dataset, with the free energy scoring~\cite{liu2020energy}. As we can see, the centre area of the learned watermark largely preserves the original input pattern, containing the semantic message that guides the detection primitively. By contrast, the edge area of the original input is superimposed by the specific pattern of the watermark, which may \emph{encode} the knowledge once hidden by the model in boosting OOD detection. Overall, watermarking can preserve the meaningful pattern of original inputs in detection, with the improved detection capability that is learned from the trained model and ID data. 

Figure~\ref{fig:kde} demonstrates the effect of our learned watermark, which is an example with the free energy scoring. After watermarking, the scoring distributions are much concentrated, and the gap between ID (i.e., CIFAR-$10$) and OOD (i.e., SVHN~\cite{netzer2011reading} and Texture~\cite{cimpoi2014describing} datasets) data is enlarged notably. We conduct extensive experiments for a wide range of OOD evaluation benchmarks 
, and the results verify the effectiveness of our proposal. 

The success of watermarking takes roots in the following aspects: (1) a well-trained model on classification has the potential to be reprogrammed for OOD detection since they are two related tasks; (2) reprogramming has been widely studied, ranging from image classification to time series analysis~\cite{deng2009imagenet,TsaiCH20}, making our proposal general across various domains; and (3) OOD detection suffers from the lack of knowledge about the real-world OOD distributions. Fortunately, with only data-level manipulation in low dimensions, watermarking can largely mitigate this issue of limited data. Overall, this data-level manipulation is orthogonal to existing methods, and thus provides a new road in OOD detection and can inspire more ways to design OOD detection methods in the future.

\section{Related Works}

To begin with, we briefly review the related works in OOD detection and model reprogramming. Please refer to Appendix~\ref{app:related work} for the detailed discussion. 

\textbf{OOD Detection} discerns ID and OOD data by their gaps regarding the specified metrics/scores, and existing methods can be roughly divided into three categories~\cite{yang2021generalized}, the \emph{classification-based} methods, the \emph{density-based} methods, and the \emph{distance-based} methods. Specifically, the classification-based methods~\cite{hendrycks2016baseline,huang2021importance,liu2020energy,sastry2019detecting} use representations extracted from the well-trained models in OOD scoring; and the distance-based methods~\cite{ChenLSZ20,huang2020feature,zaeemzadeh2021out} measure the distance of inputs from class centers in the embedding space. Moreover, the density-based methods estimate input density with probabilistic models~\cite{lee2018simple,RenLFSPDDL19,SerraAGSNL20}, identifying those OOD data with small likelihood values. Distance-based and density-based methods may suffer from complexity in computation~\cite{lee2018simple} and difficulty in optimization~\cite{ZhangGR21}. Therefore, more researchers focus on developing classification-based methods and have made big progress on benchmark datasets recently~\cite{huang2021importance,liu2020energy}.



\textbf{Model Reprogramming} repurposes well-trained models for new tasks with only data-level manipulation~\cite{elsayed2018adversarial}, indicating that deep models are competent for different jobs without changing any model parameter. In previous works, the data-level manipulation typically refers to a static padding pattern (different from our proposal) learned for the target task, which is added to the test-time data. The effectiveness of the model reprogramming is verified across image classification~\cite{elsayed2018adversarial,TsaiCH20} and time-series analysis~\cite{hambardzumyan2021warp,YangTC21}. In this paper, we use the reprogramming property of deep models for effective OOD detection, which has been overseen previously. 

\section{Preliminary} \label{sec: related work}

Let $\mathcal{X}\subset\mathbb{R}^d$ be the input space and $\mathcal{Y}=\{1,\ldots, c\}$ be the label space. We consider the ID distribution $D^\text{ID}_{\mathcal{X}, \mathcal{Y}}$ defined over $\mathcal{X}\times\mathcal{Y}$, the training sample $S_n=\{(\boldsymbol{x}_i,y_i)\}_{i=1}^n$ of size $n$ independently drawn from $D^\text{ID}_{\mathcal{X}, \mathcal{Y}}$, and a classification model ${f}:\mathcal{X}\rightarrow\mathbb{R}^c$ (with logit outputs) well-trained on $S_n$. 

Based on the model $f(\cdot)$, the goal of the classification-based OOD detection is to design a detection model $g:\mathcal{X}\rightarrow\{0,1\}$ that can distinguish test-time inputs 
with the ID distribution $D^\text{ID}_{\mathcal{X}}$ from those with the OOD distribution $D^\text{OOD}_{\mathcal{X}}$. In general, $D^\text{OOD}_{\mathcal{X}}$ is defined as an irrelevant distribution of which the label set has no intersection with $\mathcal{Y}$, and thus should not be predicted by $f(\cdot)$. 
Overall, with $0$ denoting the OOD case and $1$ the ID case, the detection model $g(\cdot)$ is defined as 
\begin{equation}
    g(\boldsymbol{x};\tau)=\begin{cases} 1 & {s}(\boldsymbol{x};f) \ge \tau \\ 0 & {s}(\boldsymbol{x};f) < \tau \end{cases}, \label{eq: tau}
\end{equation}
where $\tau\in\mathbb{R}$ is a threshold and $s:\mathcal{X}\rightarrow \mathbb{R}$ is the scoring function defined by $f(\cdot)$ whose parameters are fixed. Here, we focus on two representative methods in the classification-based OOD detection, namely, the \emph{softmax scoring} and the \emph{free energy scoring}. 

\textbf{Softmax Scoring Function}~\cite{hendrycks2016baseline} uses the maximum softmax prediction in OOD detection, of which the scoring function $s_\text{SM}(\cdot)$ is given by
\begin{equation}
    s_\text{SM}(\boldsymbol{x};f) = \max_k~\texttt{softmax}_k~f(\boldsymbol{x}),  \label{eq: softmax score}
\end{equation}
where $\texttt{softmax}_k (\cdot)$ denotes the $k$-th element of the softmax outputs. In general, with a large (small) $s_\text{SM}(\boldsymbol{x};f)$, the detection model will take the input $\boldsymbol{x}$ as an ID (OOD) case. 

\textbf{Free Energy Scoring Function}~\cite{liu2020energy} adopts the free energy function for scoring, defined by the logit outputs with the $\texttt{logsumexp}$ operation, namely,
\begin{equation}
    {s_\text{FE}(\boldsymbol{x};f) = \log \sum_k \exp {f_k(\boldsymbol{x})}/{T},} \label{eq: free energy score}
\end{equation}
where $T>0$ is the temperature parameter, fixed to $1$ \cite{liu2020energy}. It aligns with the density of inputs to some extent, and thus is less susceptible to the overconfidence issue than the softmax scoring~\cite{liu2020energy}. 

\section{Watermarking Strategy}
\label{sec: heuristic}

This section introduces the key concepts of watermarking for classification-based OOD detection.

\textbf{Definition.} A watermark $\boldsymbol{w}\in\mathbb{R}^d$ is a unified pattern with the exact shape as original inputs. It is added to test-time inputs statically, and we refer $\boldsymbol{w}+\boldsymbol{x}$ a \emph{watermarked input} for $\forall\boldsymbol{x}\in\mathcal{X}$. In expectation, regarding the specified scoring function $s(\cdot)$, our watermarking should make the model excel at OOD detection for watermarked data. 


\textbf{Learning Strategy.} Given the scoring function $s(\cdot)$, it is challenging to devise the exact watermark pattern by predefined rules. Therefore, for the proper watermarks in OOD detection, we need to devise learning objectives with respect to watermarks, which consider both ID and OOD data. 

{We generally have no information about the OOD distribution $D_{\mathcal{X},\mathcal{Y}}^\text{OOD}$, while we still want the model excels in discerning ID and OOD data from scoring. For this challenge, we make the model produce high scores if watermarked ID data are observed; meanwhile, we regularize the watermark such that the model will return low scores when ID patterns do not exist. From the lens of our model, the scores should remain low if a watermarked OOD input is given since the watermark is not trained to perceive OOD data, of which the patterns are very different from the ID data.}





\textbf{Benefits of Watermarking.} Watermarking directly reprograms the model to make an adaptation to our specified task of scoring, such that the detection capability of the original model is largely improved. By contrast, previous methods typically adapt to their specified tasks by only the threshold $\tau$ as in Eq.~\eqref{eq: tau}. However, it requires the trade-off between false positive (ID) and false negative (OOD) rates when densities of scoring are non-separable (cf., Figure~\ref{fig:kde}(a)). 

Further, watermarking enjoys the benefits of previous classification-based methods in that we do not modify the original training procedure in classification, making our proposal easy to be deployed in real-world systems. Although the watermark also should be learned, the parameter space is in low dimension, and the learning procedure could be conducted \emph{post-hoc} after the systems are deployed.


\textbf{Comparison with Existing Works}. In OOD detection, this paper is a first attempt in using the reprogramming property of deep models, leading to an effective learning framework named watermarking. At first glance, our methodology is seemingly similar to ODIN~\cite{LiangLS18}, which also conducts data-level perturbation for OOD detection. However, their instance-specified perturbation relies on extra backward-forward iterations during the test, which is not required in our method. Further, ODIN is designed for the softmax scoring, but our proposal is much general in OOD detection.

\section{Realizations of Watermarking Strategy} 

\label{sec: realization}

In this section, we discuss our learning framework of watermarking in detail. 

\textbf{Learning Objectives.}
As mentioned above, we need to consider the ID and OOD situations separately, with the associated loss functions denoted by $\ell^\text{ID}(\cdot)$ and $\ell^\text{OOD}(\cdot)$. For the ID case, the ID training data are required, where we make the high scores for their watermarked counterparts. By contrast, since we typically lack knowledge about the test-time OOD data, only the watermark is used here, and we expect the model to produce the score as low as possible when only perceiving the watermark.

Further, since only the watermark is adopted for training in the OOD case, the learned watermark is pretty sensitive regarding the detection model, i.e., the model may return different predictions when facing small perturbations. Thus, 
the watermarked OOD inputs may not guarantee the low scores. To this end, the watermark is further perturbed during training. Here we adopt the Gaussian noise, leading to the perturbed watermark of the form $\boldsymbol{\epsilon} + \boldsymbol{w}$ with $\boldsymbol{\epsilon}\sim\mathcal{N}(\boldsymbol{0},\sigma_1 \mathbf{I}_d)$  the \emph{independent and identically distributed} (i.i.d.) Gaussian noise of $d$-dimension (the mean $\boldsymbol{0}$ and the standard deviation $\sigma_1\mathbf{I}_d$). Then, the overall risk can be written as,
\begin{align}
    \mathcal{L}_n(\boldsymbol{w})=\underbrace{\sum_n \ell^\text{ID}(\boldsymbol{x}_i + \boldsymbol{w},y_i;f)}_{\mathcal{L}^\text{ID}_n(\boldsymbol{w})} + \beta \underbrace{\sum_n \ell^\text{OOD}(\boldsymbol{\epsilon}_j + \boldsymbol{w};f)}_{\mathcal{L}^\text{OOD}_n(\boldsymbol{w})}, \label{eq: objective}
\end{align}
with $\beta\ge 0$ the trade-off parameter, $\mathcal{L}^\text{ID}_n(\boldsymbol{w})$ the risk for ID data, and $\mathcal{L}^\text{OOD}_n(\boldsymbol{w})$ the risk for OOD data.

{\textbf{Optimization.} To find the proper watermark, we use the first-order gradient update to iteratively update watermark's elements. However, data-level optimization remains difficult in deep learning, of which the results may get stuck at suboptimal points~\cite{wang2021probabilistic}. A common approach is to use the signum of first-order gradients, guiding the updating rule of the current watermark via
\begin{equation}
    \boldsymbol{w}\leftarrow \boldsymbol{w}-\alpha \texttt{sign}(\nabla_{\boldsymbol{w}} \mathcal{L}_n(\boldsymbol{w})),
\end{equation}
where $\texttt{sign}(\cdot)$ denotes the signum function and $\alpha>0$ is the step size \cite{madry2017towards}. }

Further, for generality and insensibility, we prefer the solution that lies in the neighbourhood having uniformly low loss, i.e., with a smooth loss landscape~\cite{KeskarMNST17}. Therefore, we adopt the \emph{sharpness-aware minimization} (SAM)~\cite{ForetKMN21}, an effective optimization framework in the seek of both the low loss value and the smooth loss landscape. Specifically, given the original risk $\mathcal{L}_n(\boldsymbol{w})$, the SAM problem is:
\begin{align}
    \mathcal{L}^\text{SAM}_n(w)=\max_{\lvert\lvert \boldsymbol{\kappa}\rvert\rvert_2\le\rho} \underbrace{\left[\mathcal{L}_n(\boldsymbol{w} + \boldsymbol{\kappa}) - \mathcal{L}_n(\boldsymbol{w})\right]}_\text{sharpness} + \mathcal{L}_n(\boldsymbol{w})=\max_{\lvert\lvert \boldsymbol{\kappa}\rvert\rvert_2\le\rho} \mathcal{L}_n(\boldsymbol{w} + \boldsymbol{\kappa})
\end{align}
where $\rho\ge 0$ is a constraint. For efficiency, the SAM makes the first-order Taylor expansion w.r.t. $\boldsymbol{\kappa}$ around $\boldsymbol{0}$, obtaining the approximated solution of the form~\footnote{With an abuse of notation, we denote the estimated solution in the SAM as $\boldsymbol{\kappa}$ for simplicity.}:
\begin{equation}
    \boldsymbol{\kappa} = \rho \texttt{sign} (\nabla_{\boldsymbol{w}} \mathcal{L}_n(\boldsymbol{w}))\frac{\lvert\nabla_{\boldsymbol{w}} \mathcal{L}_n(\boldsymbol{w})\rvert^{q-1}}{\left(\lvert\lvert\nabla_{\boldsymbol{w}} \mathcal{L}_n(\boldsymbol{w})\rvert\vert_q^q\right)^{1/p}},
\end{equation}
where $1/p+1/q=1$ and we set $p=q=2$ for simplicity. Therefore, the estimation form of the SAM is written as $\mathcal{L}_n(\boldsymbol{w} + \boldsymbol{\kappa})$, with corresponding updating rule of
\begin{equation}
    \boldsymbol{w}\leftarrow \boldsymbol{w}-\alpha \texttt{sign}(\nabla_{\boldsymbol{w}} \mathcal{L}_n(\boldsymbol{w}+\boldsymbol{\kappa})), \label{eq: sign_sam}
\end{equation}
yielding an efficient optimization algorithm that induces the effective watermark. 


\textbf{The Overall Algorithm.} In summary, we describe the overall learning framework. To begin with, the watermark is initialized by the i.i.d. Gaussian noise with the $\boldsymbol{0}$ mean and a small standard deviation $\sigma_2\mathbf{I}_d$, and the learning procedure consists of three stages for each updating step:
\begin{itemize}
    \item Negative sampling: a set of noise data $\epsilon$ is sampled, assuming be of the size $m$ as that of the mini-batch regarding the ID sample;
    \item Risk calculating: the risk for ID and OOD data are computed, and the overall risk is given by their sum with a trade-off parameter $\beta$ as in Eq.~\eqref{eq: objective};
    \item Watermark updating: the first-order gradient guides the pixel-level update of the watermark, using the signum of gradients and the SAM to make a reliable update as in Eq.~\eqref{eq: sign_sam}. 
\end{itemize}

The learned watermark is added to test-time inputs for OOD detection, and the detection model with the pre-defined scoring function is then deployed. Appendix~\ref{sec: app alg} summarizes our learning framework of watermarking. Moreover, two specifications of watermarking are discussed in the following.




\textbf{Two Realizations.}
Here, we focus on two representative methods in OOD detection, namely, the softmax scoring and the free energy scoring. {For other representative methods in OOD detection, please refer to Appendix~\ref{sec: app exp} for their descriptions and the experiments.}  

\textit{Softmax Scoring-based Watermarking.} Following~\cite{HendrycksMD19}, we set $\ell^\text{ID}_\text{SM}(\cdot)$ to be the cross entropy loss and  $\ell^\text{OOD}_\text{SM}(\cdot)$ to be the cross entropy regarding the uniform distribution, namely,
\begin{equation}
    \ell^\text{ID}_\text{SM}(\boldsymbol{x},y;f)=-\log \texttt{softmax}_y f(\boldsymbol{x})~~\text{and}~~\ell^\text{OOD}_\text{SM}(\boldsymbol{x};f)=-\sum_k \frac{1}{c} \log \texttt{softmax}_k f(\boldsymbol{x}), \label{eq:obj_sm}
\end{equation}
specifying the learning objectives in Eq.~\eqref{eq: objective} for the softmax scoring-based watermarking. 

\textit{Free-Energy Scoring-based Watermarking.} \cite{liu2020energy} use a set of learning objectives for model re-training with free energy scoring. However, their $\texttt{logsumexp}$ operation originating from the free energy function is difficult for optimization, posing notorious computing issues~\cite{mann2006numerically}. To this end, we drop the $\log$ operation and make the overall risk always positive by the following learning objectives:
\begin{align}{
    \ell^\text{ID}_\text{FE} (\boldsymbol{x};f)=\sum_k \exp {-f_k(\boldsymbol{x})}/{T_1}~~\text{and}~~
    \ell^\text{OOD}_\text{FE} (\boldsymbol{x};f)=\sum_k \exp {f_k(\boldsymbol{x})}/{T_2},} \label{eq:obj_eng}
\end{align}
realizing the learning objectives in Eq.~\eqref{eq: objective} for the free energy scoring.

\section{Experiments} \label{sec: experiment}

In this section, we conduct extensive experiments for watermarking in OOD detection. Specifically, we demonstrate the effectiveness of our method on a wide range of OOD evaluation benchmarks; we conduct experiments for the important hyper-parameters in our learning framework; and we provide further experiments for an improved interpretation of our proposal. 

Baselines results are achieved by our re-run of the publicly available codes. The source code of our proposal is released at  \href{https://github.com/QizhouWang/watermarking}{github.com/qizhouwang/watermarking}.
All the methods are realized by Pytorch $1.81$ with CUDA $11.1$, where we use several machines equipped with GeForce RTX $3090$ GPUs and AMD Ryzen Threadripper $3960$X Processors.

\textbf{ID and OOD Datasets}. We use CIFAR-$10$, CIFAR-$100$~\cite{krizhevsky2009learning}, and ImageNet~\cite{ILSVRC15} datasets as three ID datasets, with data pre-processing including horizontal flip and normalization. Furthermore, for the OOD datasets, we adopt several commonly-used benchmarks, including Textures~\cite{cimpoi2014describing}, SVHN~\cite{netzer2011reading}, Places$365$~\cite{ZhouLKO018}, LSUN~\cite{yu2015lsun}, and iSUN~\cite{xu2015turkergaze}. Referring to Appendix~\ref{app: hyper-parameter} for hyper-parameter settings. 

\textbf{Evaluation Metrics}. The performance in OOD detection is measured via three representative metrics, which are all threshold-independent~\cite{DavisG06}: (1) the false positive rate of OOD sample when true positive rate of ID data is at $95\%$ (FPR$95$); (2) the \emph{area under the receiver operating characteristic curve} (AUROC), interpreted as the probability that an ID input has a greater score than an OOD one; and (3) the \emph{area under the precision-recall curve} (AUPR), which further adjusts for different base rates.
 
\textbf{Configuration}. Following previous works~\cite{liu2020energy}, we employ WideResNet~\cite{zagoruyko2016wide} (WRN-40-2) as the backbone model. For the CIFAR benchmarks, the models are trained for $200$ epochs via the stochastic gradient descent, with the batch size $64$, the momentum $0.9$, and the initial learning rate $0.1$. The learning rate is divided by $10$ after $100$ and $150$ epochs.  For the ImageNet Benchmark, the model is trained for $120$ epochs via the stochastic gradient descent, with the batch size $32$, the momentum $0.9$ and the initial learning rate $0.05$. The learning rate is divided by $10$ after $60$ and $90$ epochs. 

\begin{figure*}[t]
\centering  
\subfigure[softmax scoring on CIFAR-$10$]{
\centering  
\begin{minipage}[t]{0.42\textwidth}
    \centering
	\includegraphics[width=\linewidth]{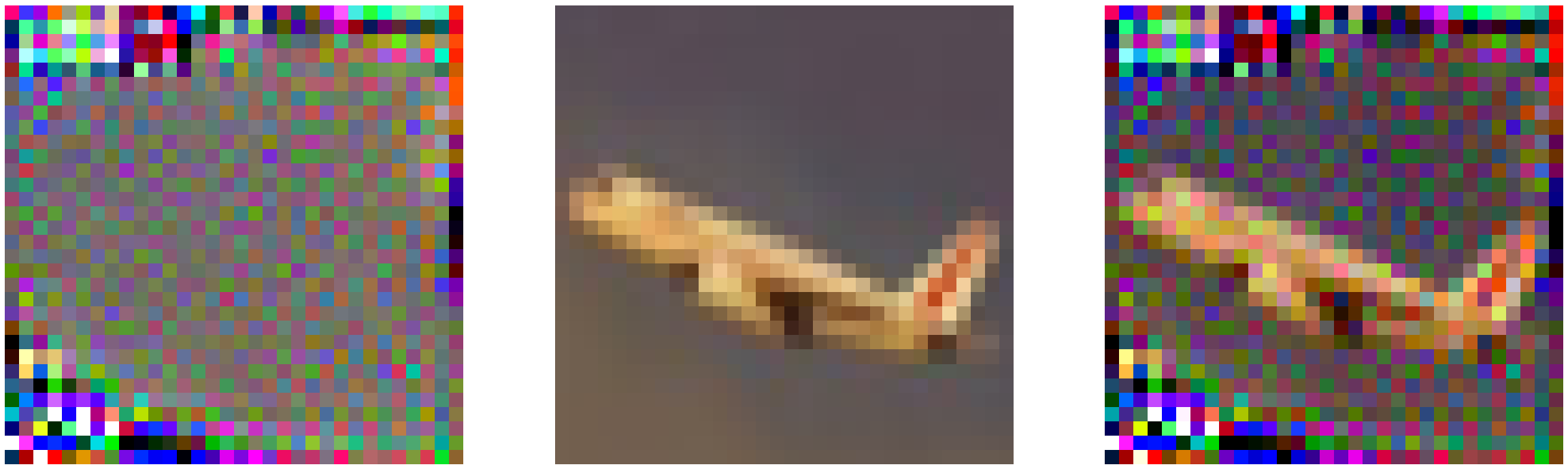}
    \centering  
\end{minipage}} 
~~~~~~~~
\subfigure[softmax scoring on CIFAR-$100$]{
\centering  
\begin{minipage}[t]{0.42\textwidth}
    \centering  
    \includegraphics[width=\linewidth]{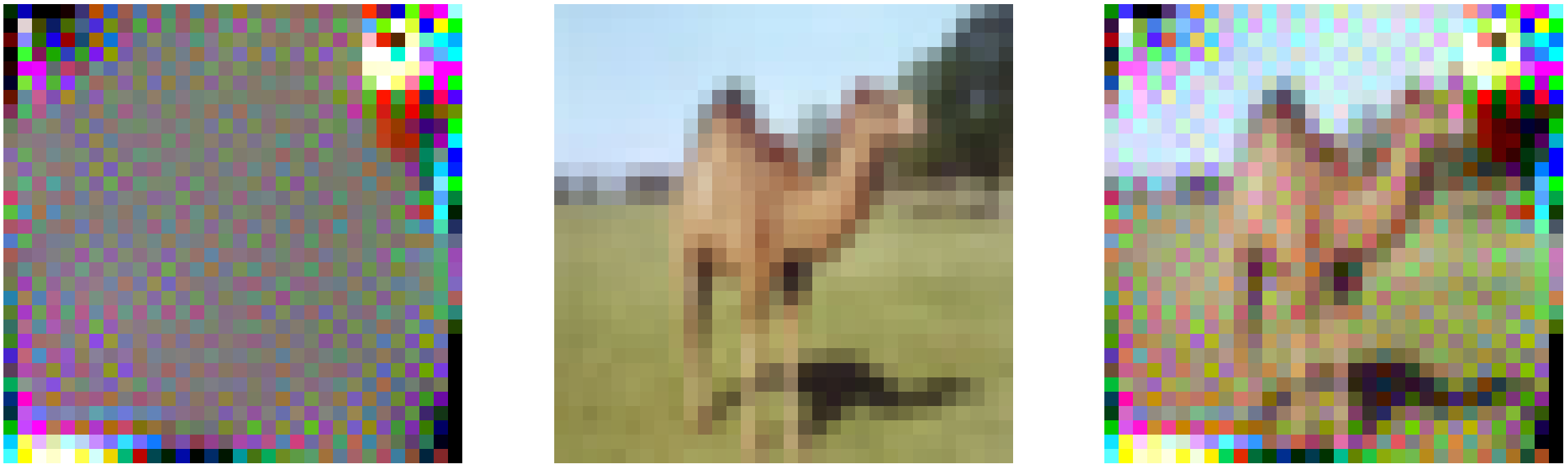}
    \centering
\end{minipage}}

\subfigure[free energy scoring on CIFAR-$10$]{
\centering  
\begin{minipage}[t]{0.42\textwidth}
    \centering  
    \includegraphics[width=\linewidth]{energy_cifar10.png}
    \centering
\end{minipage}}
~~~~~~~~
\subfigure[free energy scoring on CIFAR-$100$]{
\centering  
\begin{minipage}[t]{0.42\textwidth}
    \centering  
    \includegraphics[width=\linewidth]{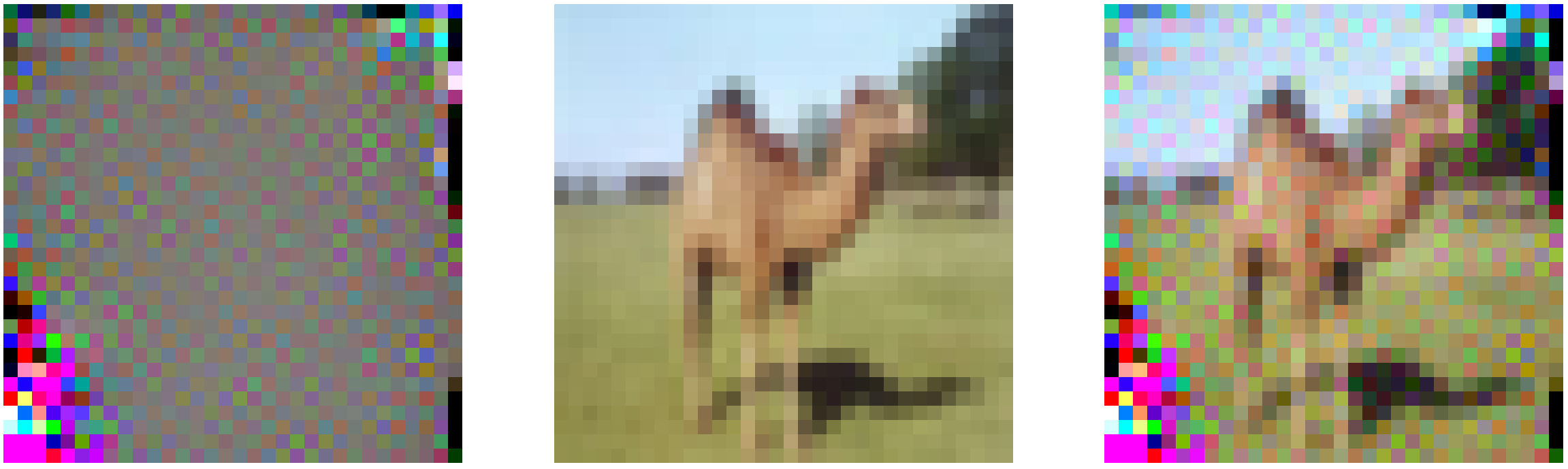}
    \centering
\end{minipage}}

\caption{The learned watermarks (left) and the example images with (middle) and without (right) the watermarks. All the pictures are clamped between $0$ and $255$ for the purpose of illustration.} 
\label{fig: learned watermark}

\end{figure*}

\textbf{CIFAR Benchmarks}. 
We depict the learned watermarks in Figure~\ref{fig: learned watermark}. As we can see, the centre areas maintain the pattern of original inputs, which are helpful in OOD detection primitively. By contrast, the edge areas of the watermarks distort the original features, superimposed with the pattern that may further boost the capability of the original models in OOD detection. 

Then, we demonstrate the improvement of watermarking on CIFAR-$10$ and CIFAR-$100$ datasets. For the results of the softmax scoring in Table~\ref{tab: softmax}, our watermarking reduces the average FPR$95$ by $2.99\sim12.84$, boosts the average AUROC by $3.09\sim3.69$ and the average AUPR by $1.10\sim1.18$. Moreover, for each of the considered test-time OOD dataset, there also exist stable improvements after watermarking, except for the comparable results regarding Places$365$ on CIFAR-$100$ dataset. 

For the results of the free energy scoring in Table~\ref{tab: free energy}, the improvement via watermarking is also substantial, with $4.54\sim13.98$, $3.43\sim4.65$, and $1.13\sim1.37$ better performance regarding the average FPR$95$, AUROC, and AUPR. Overall, Table~\ref{tab: softmax} and Table~\ref{tab: free energy} not only justify the effectiveness of watermarking and also demonstrate the generality of the proposed watermarking. Further, comparing between CIFAR-$10$ and CIFAR-$100$ datasets, the improvements of watermarking on CIFAR-$10$ is much greater than that of CIFAR-$100$, aligning with the previous observations~\cite{HuangL21} that a large semantic space can exaggerate the challenge in effective OOD detection.

\begin{table}[t]
\parbox{.47\linewidth}{
\centering
\caption{Softmax scoring with/without watermarking on CIFAR benchmarks. $\downarrow$ ($\uparrow$) indicates smaller (larger) values are preferred.} \label{tab: softmax}
\vspace{5pt}
\scriptsize
{
\begin{tabular}{c|ccc}
\toprule[1.5pt]
                   & FPR95 $\downarrow$     & AUROC $\uparrow$       & AUPR $\uparrow$      \\
                   \cline{2-4} 
\multirow{-2}{*}{} & \multicolumn{3}{c}{w/ (w/o) watermark} \\
\midrule[0.6pt]
\multicolumn{4}{c}{\cellcolor{greyL}CIFAR-10} \\
\midrule[0.6pt]
iSUN               & \textbf{43.60} (55.55) & \textbf{93.53} (90.14) & \textbf{98.67} (97.84) \\ 
Places$365$          & \textbf{60.75} (62.50) & \textbf{87.85} (87.41) & \textbf{96.98} (96.94) \\
Texture            & \textbf{42.00} (59.30) & \textbf{92.83} (88.37) & \textbf{98.43} (97.14) \\
SVHN               & \textbf{27.25} (49.10) & \textbf{96.00} (91.69) & \textbf{99.17} (96.54) \\
LSUN             & \textbf{40.70} (52.05) & \textbf{94.36} (91.50) & \textbf{98.86} (98.16) \\
\midrule
\textbf{average}   & \textbf{42.86} (55.70) & \textbf{92.91} (89.82) & \textbf{98.42} (97.32) \\ \midrule[1pt]
\multicolumn{4}{c}{\cellcolor{greyL}CIFAR-100} \\
\midrule[1pt]
iSUN               & \textbf{77.85} (83.35) & \textbf{79.91} (75.28) & \textbf{95.35} (94.00) \\ 
Places$365$          & 83.25 (\textbf{82.20}) & 74.28 (\textbf{74.40}) & \textbf{93.47} (93.44) \\
Texture            & \textbf{79.10} (83.80) & \textbf{77.14} (72.83) & \textbf{94.26} (92.81) \\
SVHN               & \textbf{82.95} (85.05) & \textbf{76.92} (70.64) & \textbf{94.72} (92.61) \\
LSUN             & \textbf{76.75} (80.45) & \textbf{79.60} (76.25) & \textbf{95.27} (94.32) \\
\midrule
\textbf{average}   & \textbf{79.98} (82.97) & \textbf{77.57} (73.88) & \textbf{94.61} (93.43) \\ \bottomrule[1.5pt]      
\end{tabular}
}}~~~~
\parbox{.47\linewidth}{
\centering
\caption{Free energy scoring with/without watermarking on CIFAR benchmarks. $\downarrow$ ($\uparrow$) indicates smaller (larger) values are preferred.} \label{tab: free energy}
\vspace{5pt}
\scriptsize{
\begin{tabular}{c|ccc}
\toprule[1.5pt]
                   & FPR$95$ $\downarrow$     & AUROC $\uparrow$       & AUPR $\uparrow$      \\
                   \cline{2-4} 
\multirow{-2}{*}{} & \multicolumn{3}{c}{w/ (w/o) watermark} \\
\midrule[0.6pt]
\multicolumn{4}{c}{\cellcolor{greyL}CIFAR-10} \\
\midrule[0.6pt]
iSUN               & \textbf{16.30} (32.10) & \textbf{96.97} (92.84) & \textbf{99.39} (98.33) \\
Places$365$          & \textbf{36.25} (41.45) & \textbf{91.87} (89.65) & \textbf{97.94} (97.21) \\
Texture            & \textbf{32.60} (52.05) & \textbf{93.14} (85.43) & \textbf{98.08} (95.52) \\
SVHN               & \textbf{16.45} (35.25) & \textbf{97.11} (90.91) & \textbf{99.39} (97.68) \\
LSUN             & \textbf{16.85} (27.50) & \textbf{96.97} (93.98) & \textbf{99.38} (98.59) \\
\midrule
\textbf{average}   & \textbf{23.69} (37.67) & \textbf{95.21} (90.56) & \textbf{98.83} (97.46) \\ \midrule[1pt]
\multicolumn{4}{c}{\cellcolor{greyL}CIFAR-100} \\
\midrule[1pt]
iSUN               & \textbf{75.05} (81.80) & \textbf{83.07} (79.04) & \textbf{96.15} (94.98) \\ 
Places$365$          & \textbf{80.45} (80.50) & \textbf{77.78} (74.99) & \textbf{94.45} (93.37) \\
Texture            & \textbf{75.15} (80.20) & \textbf{79.55} (76.00) & \textbf{94.79} (93.53) \\
SVHN               & \textbf{82.85} (85.10) & \textbf{75.26} (74.20) & \textbf{94.18} (93.70) \\
LSUN             & \textbf{71.85} (80.45) & \textbf{84.01} (78.29) & \textbf{96.33} (94.69) \\
\midrule
\textbf{average}   & \textbf{77.07} (81.61) & \textbf{79.93} (76.50) & \textbf{95.18} (94.05) \\ \bottomrule[1.5pt]      
\end{tabular}
}}
\end{table}

\begin{figure*}[t]
\centering  
\subfigure[w/o watermarking]{
\centering  
\begin{minipage}[t]{0.39\textwidth}
    \centering
	\includegraphics[width=\linewidth]{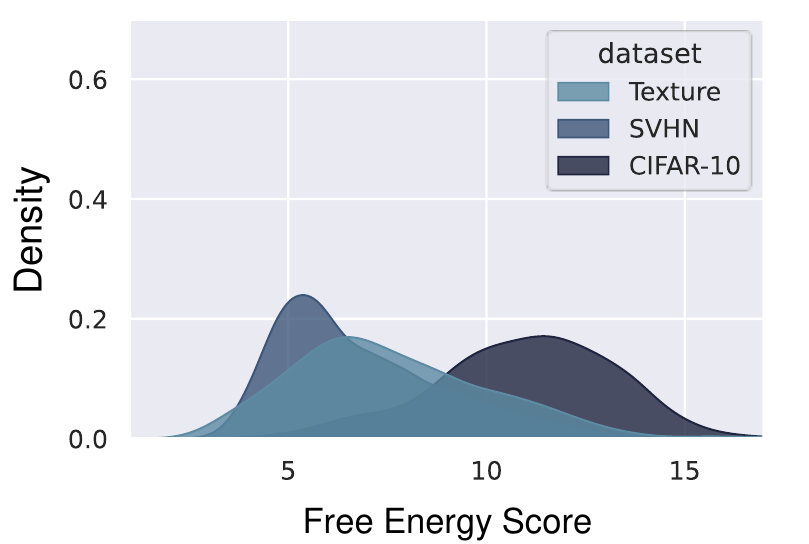}
    \centering  
\end{minipage}} 
~~~~~~~~~~~~~~~~
\subfigure[w/ watermarking]{
\centering
\begin{minipage}[t]{0.39\textwidth}
    \centering  
    \includegraphics[width=\linewidth]{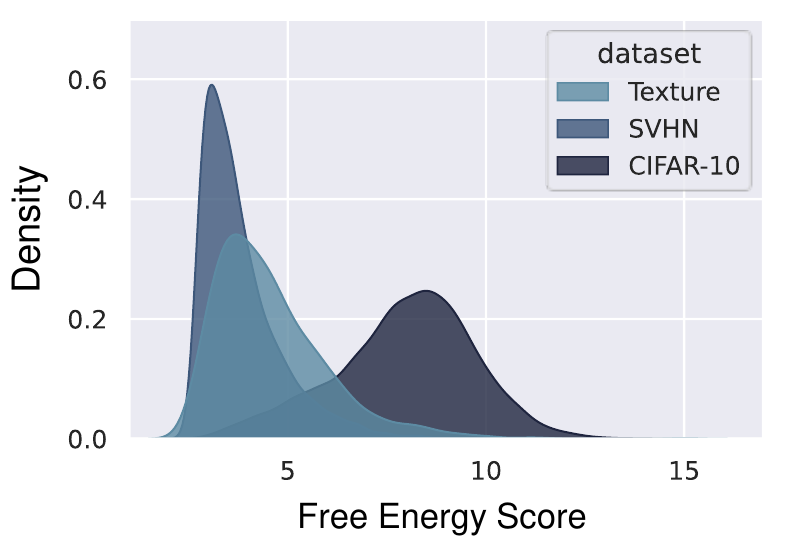}
    \centering
\end{minipage}}
\caption{An illustration on CIFAR-$10$ dataset regarding the free energy scoring. (a) depicts the scoring distributions before watermarking, and (b) is the scoring distributions after watermarking. } 
\label{fig: kde2}
\end{figure*}

Figure~\ref{fig: kde2} illustrates the scoring distributions before (a) and after (b) watermarking on CIFAR-$10$, where we take the free energy scoring as an example. Due to the space limit, we only consider two test-time OOD datasets, namely, Texture and SVHN. As we can see, after watermarking, the distribution gap between the ID (i.e., CIFAR-$10$) and the OOD (i.e., Texture and SVHN) data is enlarged, and thus the detection capability is improved.

\begin{table}[t]
\parbox{.47\linewidth}{
\centering
\caption{{The softmax scoring with/without watermarking on ImageNet. The notion $\downarrow$ ($\uparrow$) indicates smaller (larger) values are preferred.}} \label{tab: imagenet_sm}
\vspace{5pt}
\scriptsize
{
\begin{tabular}{c|ccc}
\toprule[1.5pt]
                   & FPR95 $\downarrow$     & AUROC $\uparrow$       & AUPR $\uparrow$      \\
                   \cline{2-4} 
\multirow{-2}{*}{} & \multicolumn{3}{c}{w/ (w/o) watermark} \\
\midrule[0.6pt]
iSUN               & \textbf{11.54} (52.45) & \textbf{97.41} (92.52) & \textbf{99.45} (98.70) \\
Places$365$        & \textbf{70.59} (73.25) & \textbf{82.03} (80.78) & \textbf{95.62} (94.58) \\
Texture            & \textbf{61.20} ({67.18}) & \textbf{84.00} ({82.27}) & \textbf{98.60} ({97.85}) \\
SVHN               & {44.58} (\textbf{28.49}) & {93.56} (\textbf{95.60}) & {98.70} (\textbf{99.00}) \\
LSUN               & \textbf{11.84} (54.62) & \textbf{97.97} (91.52) & \textbf{99.57} (98.25) \\
\midrule
\textbf{average}   & \textbf{40.50} (54.93) & \textbf{91.22} (88.57) & \textbf{98.42} (97.69) \\ \bottomrule[1.5pt]      
\end{tabular}
}}~~~~
\parbox{.47\linewidth}{
\centering
\caption{{The free energy scoring with/without watermarking on ImageNet. The notion $\downarrow$ ($\uparrow$) indicates smaller (larger) values are preferred.}} \label{tab: imagenet_fe}
\vspace{5pt}
\scriptsize{
\begin{tabular}{c|ccc}
\toprule[1.5pt]
                   & FPR$95$ $\downarrow$     & AUROC $\uparrow$       & AUPR $\uparrow$      \\
                   \cline{2-4} 
\multirow{-2}{*}{} & \multicolumn{3}{c}{w/ (w/o) watermark} \\
\midrule[0.6pt]
iSUN               & \textbf{32.83} (45.40) & \textbf{94.35} (94.00) & \textbf{98.89} (98.20) \\ 
Places$365$        & \textbf{71.85} (75.01) & \textbf{79.85} (78.54) & \textbf{94.65} (94.40) \\
Texture            & \textbf{67.75} (68.77) & \textbf{80.80} (80.22) & \textbf{97.00} (96.51) \\
SVHN               & \textbf{12.85} (27.60) & \textbf{97.68} (95.17) & \textbf{99.45} (99.00) \\
LSUN               & \textbf{33.75} (46.47) & \textbf{93.71} (90.59) & \textbf{98.80} (97.94) \\
\midrule
\textbf{average}   & \textbf{43.23} (52.73) & \textbf{89.10} (86.14) & \textbf{97.73} (97.15) \\ \bottomrule[1.5pt]      
\end{tabular}
}}
\end{table}

{\textbf{ImageNet Benchmark.} Huang and Li~\cite{HuangL21} show that many advanced methods developed on the CIFAR benchmarks can hardly work for the ImageNet dataset due to its large semantic space with $1$k classes. In order to verify the power of watermarking in large semantic space, we conduct experiments with ImageNet being the ID dataset, and the results regarding the softmax scoring and the free energy scoring are summarized in Table~\ref{tab: imagenet_sm} and Table~\ref{tab: imagenet_fe}. For the softmax scoring, we decrease the average FPR$95$ from $54.93$ to $40.50$ after watermarking; for the free energy scoring, we reduce the average FPR$95$ from $52.73$ to $43.23$. It demonstrates that our watermarking still works well in the case of the large semantic space, with considerable improvements in OOD detection power.}

\begin{wraptable}{r}{0.45\textwidth}
\centering
\caption{{The performance of our watermarking on near OOD detection regarding the softmax scoring and the free energy scoring. }} \label{tab: near ood}
\resizebox{.41\columnwidth}{!}{
\centering
\scriptsize{
\begin{tabular}{c|ccc}
\toprule[1.5pt]
        & FPR95 & AUROC & AUPR \\
\midrule[0.6pt]
\multicolumn{4}{c}{\cellcolor{greyL}softmax scoring} \\
\midrule[0.6pt]
w/o watermark   & 90.10 & 55.47 & 86.16 \\
\hline
common          & 88.25 & 53.16 & 84.75 \\
permute         & 86.45 & 60.04 & 86.33 \\
rotate          & \textbf{81.50} & \textbf{65.69} & \textbf{88.67} \\
\midrule[0.6pt]
\multicolumn{4}{c}{\cellcolor{greyL}free energy scoring} \\
\midrule[0.6pt]
w/o watermark   & 52.25 & 86.49 & 96.44 \\
\hline
common          & 49.75 & 88.52 & 96.98 \\
permute         & 48.55 & 88.40 & 97.03 \\
rotate          & \textbf{47.85} & \textbf{88.90} & \textbf{97.08} \\
\bottomrule[1.5pt]
\end{tabular}}}
\end{wraptable}

\textbf{Near OOD Detection.} The above experiments focus on the far OOD detection setups where ID and OOD data are different regarding semantics and styles. Now, we further demonstrate the power of our watermarking strategy in a near OOD situation~\cite{winkens2020contrastive}, covering a challenging situation where ID and OOD data have similar styles (i.e., near OOD data). Except for the common learning setup in Section~\ref{sec: realization}, we further consider the use of shifting augmentations~\cite{tack2020csi}, which are data augmentations that are harmful to the standard contrastive learning but can be used to construct near OOD data similar to ID data. We consider two representative shifting augmentations: ``permute'' (permute evenly partitioned data) and ``rotate'' (rotate 90 degrees of original data). The shifting-augmented ID data are taken as OOD data fed into  $\ell^\text{OOD}(\cdot)$ along with random Gaussian noise. 

The near OOD experiments are summarized in Table~\ref{tab: near ood}, where we take CIFAR-$10$ as ID data and CIFAR-$100$ as OOD data. Here, the common learning setup (common) already leads to improved performance compared to the cases without watermarking (w/o watermark). Moreover, watermarking with shifting augmentations (permute and rotate) can further boost the detection power of the models, leading to at most $8.60$ and $4.70$ improvements in FPR95 for the softmax and the free energy scoring.


{\textbf{Effect of Hyper-parameters.}} 
To further interpret our learning framework, we compare the performance of the learned watermarks regarding various setups of hyper-parameters, focusing on the standard deviation $\sigma_1$ in the Gaussian noise and the perturbation constraint $\rho$ in the SAM, which are both critical. As a case study, we conduct experiments regarding the softmax scoring on CIFAR-$10$. For detailed results about the ablation study, please refer to Appendix~\ref{sec: app exp}.   

 
Table~\ref{tab: ablation sigma} lists the results with various values of $\sigma_1$, ranging from $0.00$ to $2.00$. Note that $\sigma_1$ is the standard deviation of the Gaussian noise added to the watermark, with different values indicating various degrees of the perturbation. As we can see, a mild perturbation of the watermark (e.g., $\sigma_1=0.40$) can truly lead to improved results in detection, with $10.05$, $1.60$, and $0.34$ improvements regarding the average FPR$95$, AUROC, and AUPR. It indicates that the learned watermark is sensitive when facing small perturbations, if the Gaussian noise is not applied during training. However, some extreme values (e.g., $\sigma_1=2.00$) may overwhelm the watermark pattern and thus be detrimental. 

In Table~\ref{tab: ablation rho}, we summarize the experimental results given by various values of $\rho$ for the SAM. Overall, a large value of $\rho$ indicates that a wide range in the solution's neighbours should be smooth, and thus the stability of the result is expected to be improved. However, such a solution may consume too much capacity of the watermark, misleading the learning procedure to some unsatisfactory results. On the other side, our watermarking truly benefits from the SAM with a mild choice of the hyper-parameter. {Specifically, comparing with the results without the SAM (i.e., $\rho=0.00$), the detection capability of the watermark with a suitable $\rho$ (i.e., $\rho=1.0$) is largely improved, with $3.92$, $0.52$, and $0.10$ better results regarding the average FPR$95$, AUROC, and AUPR. }

\begin{table}[t]
\centering
\parbox{.47\linewidth}{
\centering
\scriptsize
\caption{The average performance of the softmax scoring on CIFAR-$10$ dataset with various values of the parameter $\sigma_1$.  The notion $\downarrow$ ($\uparrow$) indicates smaller (larger) values are preferred.} \label{tab: ablation sigma}
\resizebox{.4\columnwidth}{!}{
\vspace{5pt}
{
\begin{tabular}{c|ccc}
\toprule[1.5pt]
   $\sigma_1$          & FPR$95$ $\downarrow$     & AUROC $\uparrow$       & AUPR $\uparrow$      \\
\midrule[0.6pt]
2.00               & 42.11                  & 92.01                  & 98.24      \\
1.60               & 41.41                  & 92.14                  & 98.25       \\
1.20               & 41.98                  & 91.91                  & 98.20      \\
0.80               & 43.38                  & 91.89                  & 98.21 \\
\cellcolor{greyC}0.40               & \cellcolor{greyC}\textbf{38.66}                  & \cellcolor{greyC}\textbf{93.03}                  & \cellcolor{greyC}\textbf{98.45} \\
0.00               & 48.71                  & 91.43                  & 98.11 \\ 
\bottomrule[1.5pt]      
\end{tabular}
}}}~~~~~
\parbox{.47\linewidth}{
\centering
\caption{{The average performance of the softmax scoring on CIFAR-10 dataset with various values of the parameter $\rho$.  The notion $\downarrow$ ($\uparrow$) indicates smaller (larger) values are preferred.}} \label{tab: ablation rho}
\scriptsize
\resizebox{.4\columnwidth}{!}{
\vspace{5pt}
{
\begin{tabular}{c|ccc}
\toprule[1.5pt]
   $\rho$         & FPR$95$ $\downarrow$     & AUROC $\uparrow$       & AUPR $\uparrow$      \\
\midrule[0.6pt]
5.00               & 60.02                  & 87.36                  & 97.15       \\
\cellcolor{greyC}1.00               & \cellcolor{greyC}\textbf{39.12}                  & \cellcolor{greyC}\textbf{92.96}                  & \cellcolor{greyC}\textbf{98.42}       \\
0.50               & 43.55                  & 92.38                  & 98.34       \\ 
0.10               & 41.99                  & 92.77                  & 98.41       \\
0.05               & 42.06                  & 92.84                  & 98.42       \\ 
0.00               & 43.04                  & 92.44                  & 98.32       \\
\bottomrule[1.5pt]      
\end{tabular}
}
}}
\end{table}

\begin{wraptable}{r}{0.45\linewidth}
\centering
\caption{Transferability of watermarking across scoring functions. SM denotes softmax scoring and FE denotes free energy scoring.  } \label{tab: transferability}
\vspace{7pt}
\scriptsize{
\begin{tabular}{cc|ccc}
\toprule[1.5pt]
 learn & score & FPR$95$ $\downarrow$     & AUROC $\uparrow$       & AUPR $\uparrow$      \\
\midrule[0.6pt]
\multicolumn{5}{c}{\cellcolor{greyL}CIFAR-$10$} \\
\midrule[0.6pt]
SM & SM & 42.86 & 92.91 & 98.42 \\
FE & SM & \textbf{40.19} & \textbf{94.83} & \textbf{98.99} \\
\midrule[0.6pt]
FE & FE & \textbf{23.69} & \textbf{95.21} & \textbf{98.83} \\
SM & FE & 28.22 & 94.70 & 98.81 \\
\midrule[0.6pt]
\multicolumn{5}{c}{\cellcolor{greyL}CIFAR-$100$} \\
\midrule[0.6pt]
SM & SM & 79.98 & \textbf{77.57} & \textbf{94.61} \\
FE & SM & \textbf{77.07} & {76.27} & {94.02} \\
\midrule[0.6pt]
FE & FE & \textbf{77.07} & \textbf{79.93} & \textbf{95.18} \\
SM & FE & 78.48 & {79.24} & {94.56} \\
\bottomrule[1.5pt]      
\end{tabular}
}
\end{wraptable}

\textbf{Transferability of watermarking}. Further, we explore the transferability of the watermarking strategy learned with different scoring functions, e.g., we study the effect of the watermark learned with the softmax scoring when it is deployed with the free energy scoring. The results are summarized in Table~\ref{tab: transferability}. {Note that, a ``learn'' with FE and ``score'' with SM indicates that the watermark is learned with the free energy scoring and tested regarding the softmax scoring.} As we can see, the watermarks learned with the free energy scoring can be reused for the softmax scoring (even with better results), while the reverse leads a deterioration. It indicates that our learning objectives in Eq.~\eqref{eq:obj_eng} are general and effective for both the softmax scoring and the free energy scoring. However, we do not observe any transferability between different datasets. For example, for the softmax scoring, when the watermark learned on CIFAR-10 is adopted for CIFAR-100, there is a drop of performance with $51.93$ (from $42.86$ to $94.79$), $35.29$ (from $92.91$ to $57.62$) and $12.00$ ($98.42$ to $86.42$) regarding the average FPR$95$, AUROC, and AUPR. Further studies may be required here.

\section{Conclusion}

This paper demonstrates that the model's inherent capability in OOD detection can be largely improved with only data-level manipulation, where we propose a general and effective methodology named watermarking. Overall, we learn a discriminative pattern that could be superimposed onto original inputs, such that the OOD scores become much separable between ID and OOD data. Our results indicate a promising direction in OOD detection that warrants our exploration in the future. Further investigation should focus on the stability in optimization and general learning objectives regarding advanced OOD detection methods.

\begin{ack}
QZW, YGZ and BH were supported by the RGC Early Career Scheme No.~22200720, NSFC Young Scientists Fund No.~62006202, Guangdong Basic and Applied Basic Research Foundation No.~2022A1515011652, RGC Research Matching Grant Scheme No.~RMGS2022\_11\_02 and No.~RMGS2022\_13\_06, and HKBU CSD Departmental Incentive Grant. CG was supported by NSF of China No.~61973162, NSF of Jiangsu Province No.~BZ2021013, NSF for Distinguished Young Scholar of Jiangsu Province No.~BK20220080, and the Fundamental Research Funds for the Central Universities No.~30920032202 and No.~30921013114. TL was partially supported by Australian Research Council Projects DP180103424, DE-190101473, IC-190100031, DP-220102121, and FT-220100318.
\end{ack}

\bibliography{neurips_2022.bib}
\bibliographystyle{unsrtnat}


\section*{Checklist}

\begin{enumerate}

\item For all authors...
\begin{enumerate}
  \item Do the main claims made in the abstract and introduction accurately reflect the paper's contributions and scope?
    \answerYes{}
  \item Did you describe the limitations of your work?
    \answerYes{}
  \item Did you discuss any potential negative societal impacts of your work?
    \answerNo{}
  \item Have you read the ethics review guidelines and ensured that your paper conforms to them?
    \answerYes{}
\end{enumerate}

\item If you are including theoretical results...
\begin{enumerate}
  \item Did you state the full set of assumptions of all theoretical results?
    \answerNA{}
	\item Did you include complete proofs of all theoretical results?
    \answerNA{}
\end{enumerate}

\item If you ran experiments...
\begin{enumerate}
  \item Did you include the code, data, and instructions needed to reproduce the main experimental results (either in the supplemental material or as a URL)?
    \answerYes{}
  \item Did you specify all the training details (e.g., data splits, hyperparameters, how they were chosen)?
    \answerYes{}
	\item Did you report error bars (e.g., with respect to the random seed after running experiments multiple times)?
    \answerYes{}
	\item Did you include the total amount of compute and the type of resources used (e.g., type of GPUs, internal cluster, or cloud provider)?
    \answerYes{}
\end{enumerate}

\item If you are using existing assets (e.g., code, data, models) or curating/releasing new assets...
\begin{enumerate}
  \item If your work uses existing assets, did you cite the creators?
    \answerNA{}
  \item Did you mention the license of the assets?
    \answerNA{}
  \item Did you include any new assets either in the supplemental material or as a URL?
    \answerNo{}
  \item Did you discuss whether and how consent was obtained from people whose data you're using/curating?
    \answerNo{We use only standard datasets}
  \item Did you discuss whether the data you are using/curating contains personally identifiable information or offensive content?
    \answerNo{We use only standard datasets}
\end{enumerate}

\item If you used crowdsourcing or conducted research with human subjects...
\begin{enumerate}
  \item Did you include the full text of instructions given to participants and screenshots, if applicable?
    \answerNA{}
  \item Did you describe any potential participant risks, with links to Institutional Review Board (IRB) approvals, if applicable?
    \answerNA{}
  \item Did you include the estimated hourly wage paid to participants and the total amount spent on participant compensation?
    \answerNA{}
\end{enumerate}

\end{enumerate}


\appendix
\onecolumn

\section{Additional Related Works} \label{app:related work}

We review the recent studies in OOD detection, model reprogramming, and backdoor attack. 

\subsection{OOD Detection}

Following~\cite{yang2021generalized}, we attribute existing works into three categories, namely, the \emph{classification-based} methods, the \emph{density-based} methods, and the \emph{distance-based} methods. In general, these methods aim to maximize the gap between ID and OOD data regarding specified metrics in identifying OOD data. 

\textbf{The classification-based methods} use the representations extracted from the well-trained classification models in OOD scoring. For example, \cite{hendrycks2016baseline,LiangLS18,liu2020energy,morteza2021provable,sun2021react,wang2021can} employ logit outputs from models in estimating the confidence of ID data; \cite{lee2018simple,sastry2019detecting} adopt Mahalanobis distance and Gram Matrix to exploit models' detection capability from embedding features; \cite{huang2021importance,LiangLS18} further demonstrate the importance of gradient information, either perturbing inputs with its gradients or directly using the gradient norm in scoring. The classification-based methods are easy to be deployed without modifying the models~\cite{yang2021generalized}, and thus it is the main focus in this paper. 

\textbf{The distance-based methods} measure the distance regarding the embedding space, taking those data far away from the class prototypes as the OOD data. Representative works adopt the Mahalanobis distance~\cite{huang2021importance,lee2018simple}, the cosine similarity~\cite{ChenLSZ20,zaeemzadeh2021out}, and the Euclidean distance~\cite{huang2020feature}. Our methods can also be used in the distance-based methods. However, extra computation, such as calculating the precision matrix~\cite{lee2018simple}, may lead difficulty in devising proper learning objectives, out of the scope of our paper. 

\textbf{The density-based methods} explicitly estimate the density of ID samples with various probabilistic models, identifying those OOD data based on the likelihood~\cite{lee2018simple}, the likelihood ratio~\cite{li2022ood,RenLFSPDDL19,SerraAGSNL20}, and the likelihood regret~\cite{XiaoYA20}. Typically, the input density is modelled by the mixture of Gaussian models~\cite{lee2018simple} and the flow-based methods~\cite{KirichenkoIW20,nalisnickdo19}. Although the density-based methods can directly characterize the properties of ID density, these methods are difficult to be trained and may make overconfident predictions, as demonstrated in previous works~\cite{morteza2021provable,nalisnickdo19}. 

Recent works also focus on the causes of challenges in OOD detection, from the lens of the BatchNorm statistics~\cite{sun2021react}, the density estimation~\cite{morteza2021provable}, and the spurious correlation~\cite{sastry2019detecting}. Improved methods, related to specified model architectures~\cite{devries2018learning,wang2021energy}, data perturbation~\cite{BitterwolfM020,ChoiC20,LiangLS18}, data augmentation~\cite{0001AB19,HendrycksMCZGL20,ThulasidasanCBB19}, and outlier exposure~\cite{HendrycksMD19,huang2022harnessing}, are also well-studied. However, these methods typically overlook the reprogramming property of deep models, which remain orthogonal to our proposal.


\subsection{Reprogramming Property}

The seminal work~\cite{elsayed2018adversarial} introduces adversarial reprogramming as an attack method in adversarial learning~\cite{GoodfellowSS14}, adversarially reprogramming the target model to perform a new task without changing the original model. The term ``attack'' lies in the fact that, by reprogramming, an attacker can easily steal public machine learning services, abusing their computational resources for tasks that violate their original purposes. Overall, \cite{elsayed2018adversarial} claim the reprogramming property of deep models---without modifying parameters of a well-trained model, we can reprogram it for a new purpose with only data-level manipulation. The reprogramming property of deep learning is preliminarily verified for various tasks, and its further applications are not limited to adversarial learning. Actually, advanced works often take reprogramming as an effective transfer learning technique in the cases with limited data and computing resources. In the context of the image classification, \cite{elsayed2018adversarial} reprogram a model trained on ImageNet in solving vision-based counting tasks; and \cite{TsaiCH20} further consider reprogramming a black-box system for biomedical image classification, which suffers from label scarcity issue. In the context of the natural language processing, well-trained models are reprogrammed for time-series classification~\cite{YangTC21} and sentiment analysis~\cite{hambardzumyan2021warp}, where data scarcity issues frequently occur. 

In this paper, we also employ the reprogramming property of deep models for transfer learning. However, instead of reprogramming across different datasets, we reprogram our original classification model for the task in OOD detection, considering the situation with the same (ID) dataset before/after reprogramming. Further, to preserve the benefits of previous classification-based detection methods, we adopt the perturbing pattern (i.e., the watermark) on the same shape as the original inputs, instead of reshaping original inputs and adding padding features as previous works~\cite{elsayed2018adversarial,YangTC21}. 

\subsection{Backdoor Attack}

Model reprogramming is also related to the \emph{backdoor attack}, which also change models' behaviour during the test. Overall, when a backdoor is \emph{embedded} during training and the \emph{trigger} is activated during the test, model predictions will be modified to the \emph{attacker-specified} labels deliberately~\cite{GuLDG19}. Nowadays, \emph{data poisoning}~\cite{GuLDG19,li2021invisible,LiuM0020,SahaSP20} is among the best to realize the backdoor attack for deep models---a portion of the training sample is modified with the attacker-specified pattern (i.e., pre-defined trigger) and the attacker-specified labels. The target models are trained on these poisoned data, and the resultant models will suffer from the backdoor attack when the trigger is activated. Please refer to~\cite{li2020backdoor} for a comprehensive survey. 

However, the backdoor attack and the model reprogramming exploit different aspects of deep models. In general, the backdoor attack utilises the \emph{excessive} learning ability in memorizing noise features~\cite{li2020backdoor}, while the reprogramming property states that the well-trained models can be reprogrammed for new tasks without modifying the original models. 

\section{The Overall Algorithm} \label{sec: app alg}

The overall learning framework is summarized in Algorithm~\ref{alg: wm}, optimizing in a stochastic manner with \texttt{num\_step} iterations. The watermark is initialized by the Gaussian noise with the $\boldsymbol{0}$ mean and a small standard deviation $\sigma_2\mathbf{I}_d$ (Step~$2$), and the learning procedure consists of three stages for each update: (1) a set of Gaussian noise data is sampled, assuming be of the size $m$ as that of the mini-batch regarding the ID sample (Step~$4$); (2) the risk for ID and OOD data are computed and the overall risk is given by their sum with a trade-off parameter $\beta$ (Step~$5$); (3) the first-order gradient guides the pixel-level update of the watermark, using the signum of gradients and the SAM to make a reliable update (Step~$7$). After watermark training, the learned watermark is added to test-time inputs for OOD detection and the detection model with the pre-defined scoring function is deployed.

\begin{algorithm}[t]  
\caption{\textbf{Watermarking} --- the learning framework.} 
\label{alg: wm}
\begin{algorithmic}[1]
\State {\bfseries Inputs:} trained model $f(\cdot)$ and ID training set $S_n$;
\State $\boldsymbol{w}\sim\mathcal{N}(\boldsymbol{0},\sigma_2 \mathbf{I}_d)$; \Comment{\textcolor{blue}{watermark initialization}}
\For{$t=1$ \textbf{to} \texttt{num\_step}}
    \State mini-batch $\{(\boldsymbol{x}_i, y_i)\}_{i=1}^m$ and noise $\{\boldsymbol{\epsilon}_i\}_{i=1}^m$; \Comment{\textcolor{blue}{data sampling}}
    \State $\mathcal{L}_m(\boldsymbol{w})=\mathcal{L}_m^\text{ID}(\boldsymbol{w})+\beta \mathcal{L}_m^\text{OOD}(\boldsymbol{w})$;
    \Comment{\textcolor{blue}{risk calculation}}
     \State $\nabla_{\boldsymbol{w}}\mathcal{L}_m(\boldsymbol{w})$ and $\boldsymbol{\kappa}$;
    \Comment{\textcolor{blue}{gradient calculation}}
    \State $\boldsymbol{w}\leftarrow \boldsymbol{w} - \alpha \nabla_{\boldsymbol{w}}\mathcal{L}_m(\boldsymbol{w})\lvert_{\boldsymbol{w}+\boldsymbol{\kappa}}$;
    \Comment{\textcolor{blue}{watermark updating}}
\EndFor
\State {\bfseries Output:} learned watermark $\boldsymbol{w}$. 
\end{algorithmic}
 \end{algorithm}

\section{Further Experiments} \label{sec: app exp}

This section conducts further experiments about our proposal.

\subsection{Impact on Test Accuracy}

\begin{table}[t]
\centering
\caption{{Test accuracy before/after watermarking. w/o denotes the benchmark without watermarking, SE (FE) denotes the softmax (free energy) scoring with watermarking.}} \label{tab: accu}
\vspace{5pt}
\small
\begin{tabular}{c|ccc||c|ccc}
\toprule[1.5pt]
 dataset & {w/o} & {SM} & {FE} & dataset & {w/o} & {SM} & {FE} \\
\midrule[0.6pt]
CIFAR-10             & \textbf{94.84\%}                   & 91.85\%                  & 93.49\%                  & CIFAR-100             & \textbf{75.96\%}                   & 72.03\%                  & 74.08\%                  \\  
\bottomrule[1.5pt]    
\end{tabular}
\end{table}

{To begin with, we study the impact of watermarking on the classification accuracy in Table~\ref{tab: accu}, comparing with the results without watermarking.  As we can see, watermarking has a negative impact on the test accuracy, dropping from $94.84\%$ to $91.85\%$ and $93.49\%$ on CIFAR-$10$; and from $75.96\%$ to $72.03\%$ and $74.08\%$ on CIFAR-$100$. Further, after watermarking, the classification accuracy with the free energy scoring is much better than that of the softmax scoring, with only $1.35\%$ to $1.88\%$ decrease in classification accuracy. Therefore, we suggest using the free energy scoring in watermarking as a default setup, which leads to better detection capability and largely preserves the original capability in classification.}

\subsection{Other Scoring Strategies with Watermarking}

\begin{table}[t]
\parbox{.47\linewidth}{
    \centering
\caption{OE with/without watermarking  on CIFAR benchmarks. $\downarrow$ ($\uparrow$) indicates smaller (larger) values are preferred.} \label{tab: oe softmax}
\vspace{5pt}
\scriptsize
{
\begin{tabular}{c|ccc}
\toprule[1.5pt]
                   & FPR95 $\downarrow$     & AUROC $\uparrow$       & AUPR $\uparrow$      \\
                   \cline{2-4} 
\multirow{-2}{*}{} & \multicolumn{3}{c}{w/ (w/o) watermark} \\
\midrule[0.6pt]
\multicolumn{4}{c}{\cellcolor{greyL}CIFAR-10} \\
\midrule[0.6pt]
iSUN               & \textbf{2.70} (2.75) & \textbf{99.54} (99.55) & \textbf{99.91} (99.90) \\ 
Places$365$        & \textbf{35.05} (37.75) & \textbf{94.22} (93.12) & \textbf{98.70} (98.39) \\ 
Texture            & \textbf{25.80} (27.95) & \textbf{95.58} (95.30) & \textbf{98.99} (98.92) \\ 
SVHN               & \textbf{30.25} (35.85) & \textbf{95.40} (94.31) & \textbf{99.83} (99.04) \\ 
LSUN               & {1.50} (\textbf{0.50}) & {99.71} (\textbf{99.81}) & \textbf{99.94} (99.94) \\ 
\midrule
\textbf{average}   & \textbf{19.06} (20.96) & \textbf{96.89} (96.41) & \textbf{99.39} (99.31) \\  \midrule[1pt]
\multicolumn{4}{c}{\cellcolor{greyL}CIFAR-100} \\
\midrule[1pt]
iSUN               & \textbf{22.85} (40.55) & \textbf{95.27} (90.60) & \textbf{98.92} (97.84) \\ 
Places$365$        & \textbf{71.75} (73.75) & \textbf{78.60} (77.89) & \textbf{94.66} (94.48) \\ 
Texture            & \textbf{66.70} (68.20) & {81.30} (\textbf{81.50}) & {95.44} (\textbf{95.49}) \\ 
SVHN               & {89.95} (\textbf{84.80}) & {68.20} (\textbf{72.40}) & {92.24} (\textbf{93.45}) \\ 
LSUN               & \textbf{19.75} (34.20) & \textbf{96.17} (92.56) & \textbf{99.15} (98.32) \\ 
\midrule
\textbf{average}   & \textbf{54.20} (60.30) & \textbf{83.90} (82.99) & \textbf{96.08} (95.91) \\  \bottomrule[1.5pt]      
\end{tabular}
}}~~~~
\parbox{.47\linewidth}{
    \centering
\caption{{MaxLogit with/without watermarking  on CIFAR benchmarks. $\downarrow$ ($\uparrow$) indicates smaller (larger) values are preferred.}} \label{tab: maxlogit}
\vspace{5pt}
\scriptsize
{
\begin{tabular}{c|ccc}
\toprule[1.5pt]
                   & FPR95 $\downarrow$     & AUROC $\uparrow$       & AUPR $\uparrow$      \\
                   \cline{2-4} 
\multirow{-2}{*}{} & \multicolumn{3}{c}{w/ (w/o) watermark} \\
\midrule[0.6pt]
\multicolumn{4}{c}{\cellcolor{greyL}CIFAR-10} \\
\midrule[0.6pt]
iSUN               & \textbf{24.40} (34.90) & \textbf{96.07} (92.54) & \textbf{99.21} (98.28) \\ 
Places$365$        & 50.60 (\textbf{43.70}) & 89.03 (\textbf{89.41}) & 97.28 (\textbf{96.81}) \\ 
Texture            & \textbf{31.35} (51.10) & \textbf{93.72} (85.88) & \textbf{98.36} (95.72) \\ 
SVHN               & \textbf{18.80} (35.70) & \textbf{96.81} (91.25) & \textbf{99.33} (97.75) \\ 
LSUN               & \textbf{22.65} (29.10) & \textbf{96.28} (93.73) & \textbf{99.23} (98.63) \\ 
\midrule
\textbf{average}   & \textbf{29.56} (38.90) & \textbf{94.38} (90.40) & \textbf{98.68} (97.91) \\  \midrule[1pt]
\multicolumn{4}{c}{\cellcolor{greyL}CIFAR-100} \\
\midrule[1pt]
iSUN               & \textbf{76.20} (78.45) & \textbf{83.18} (79.50) & \textbf{96.23} (95.06) \\ 
Places$365$        & \textbf{79.20} (80.55) & \textbf{77.19} (75.24) & \textbf{94.20} (93.42) \\ 
Texture            & \textbf{67.75} (78.00) & \textbf{83.89} (77.15) & \textbf{96.14} (93.89) \\ 
SVHN               & \textbf{81.10} (84.00) & \textbf{80.75} (73.66) & \textbf{95.79} (93.53) \\ 
LSUN               & \textbf{72.60} (78.35) & \textbf{84.20} (78.95) & \textbf{96.49} (94.84) \\ 
\midrule
\textbf{average}   & \textbf{75.37} (79.87) & \textbf{81.85} (76.90) & \textbf{95.77} (94.15) \\  \bottomrule[1.5pt]      
\end{tabular}
}}
\end{table}

Note that watermarking is orthogonal to much of the existing methods and the watermarking strategy can boost many other advanced OOD detection methods. To further verify the generality and the effectiveness of our proposal, we utilize watermarking for three representative OOD detection methods, namely, OE~\cite{HendrycksMD19}, MaxLogit~\cite{hendrycks2021improving}, ODIN~\cite{LiangLS18}, and ReAct~\cite{sun2021react}. 

For OE, Hendrycks et al.~\cite{HendrycksMD19} state that the target model can benefit from fine-tuning with extra OOD training data. In general, OE requires to re-train the target model, which will be prohibitively expensive for many real-world applications. However, since the model has seen some kinds of OOD data during training, it typically reveals superior results than many other advanced detection methods.

We are interested in whether our watermarking can improve the detection capability of the models that have been fine-tuned with OE. Typically, we assume that the training-time OOD data are different from that of the test time. Therefore, following previous works~\cite{HendrycksMD19,liu2020energy}, we adopt the tiny-ImageNet~\cite{le2015tiny} as the training-time OOD data for model fine-tuning. The learning objectives regarding the model parameters in OE is similar to Eq.~\eqref{eq:obj_sm} and we follow the default hyper-parameter setups as in~\cite{HendrycksMD19}. We learn the watermarks for the fixed OE-trained models with the softmax scoring, and the results on CIFAR benchmarks are summarized in Table~\ref{tab: oe softmax}. Overall, the experimental results suggest that our watermarking can still benefit OE in effective OOD detection.

MaxLogit, ODIN, and ReAct can be viewed as the improved versions of the softmax scoring. Specifically, MaxLogit takes the maximal logit outputs in OOD scoring, which is better than softmax scoring when facing large-class setting; ODIN clamps embedding features from the second-last layer of model outputs, which can attenuate the overconfidence issue caused by the out-sized activation of abnormal hidden units; and ReAct observes that temperature scaling and adversarial feature perturbation can improve model capability in discerning OOD data from ID data. 

{For MaxLogit, we directly use the learning objective of the softmax scoring-based watermarking, with the corresponding scoring function of the form:}
\begin{equation}
    {s_\text{MaxLogit}(\boldsymbol{x};f) = \max_k~f_k(\boldsymbol{{x}}),}  \label{eq: maxlogit}
\end{equation}
{which directly use the logit outputs (instead of softmax outputs) in discerning ID and OOD data. }

Moreover, for ODIN, the associated scoring function is given by
\begin{equation}
    s_\text{ODIN}(\boldsymbol{x};f) = \max_k~\texttt{softmax}_k~f(\boldsymbol{\tilde{x}}) / T,  \label{eq: odin}
\end{equation}
where $\boldsymbol{\tilde{x}}=\boldsymbol{{x}}-{\xi}\texttt{sign}(-\nabla_{\boldsymbol{x}}\log \texttt{softmax}_y f(\boldsymbol{x}))$ is the perturbed data point and ${\xi}$ is the perturbation magnitude. For the watermark training, the learning objectives are of the form:
\begin{equation}
    \ell^\text{ID}_\text{SM}(\boldsymbol{x},y;f)=-\log \texttt{softmax}_y f(\boldsymbol{\tilde{x}})~~\text{and}~~\ell^\text{OOD}_\text{SM}(\boldsymbol{x};f)=-\sum_k \frac{1}{c} \log \texttt{softmax}_k f(\boldsymbol{\tilde{x}}),
\end{equation}
following the same definition of $\boldsymbol{\tilde{x}}$ as in Eq.~\eqref{eq: odin}.

For ReAct, we assume the feature extractor defined by the second-last of model outputs by $f_\texttt{FEA}$ and the above classifier by $f_\texttt{CLA}$, i.e., $f(\boldsymbol{x})=f_\texttt{CLA}(f_\texttt{FEA}(x))$. Then, the rectified model output is
\begin{equation}
    f_\texttt{ReAct}(\boldsymbol{x})=f_\texttt{CLA}(\min(f_\texttt{FEA}(\boldsymbol{x}),\tau)),
\end{equation}
truncating values of hidden units from the second-last layer that are above $\tau$. The corresponding learning objectives in watermark training can be written as
\begin{equation}
    \ell^\text{ID}_\text{SM}(\boldsymbol{x},y;f)=-\log \texttt{softmax}_y f_\texttt{ReAct}(\boldsymbol{{x}})~~\text{and}~~\ell^\text{OOD}_\text{SM}(\boldsymbol{x};f)=-\sum_k \frac{1}{c} \log \texttt{softmax}_k f_\texttt{ReAct}(\boldsymbol{{x}}),
\end{equation}
which is similar to that of the softmax scoring in Eq.~\eqref{eq:obj_sm}.

Tables~\ref{tab: maxlogit} to~\ref{tab: odin} summarize the experimental results on CIFAR-$10$ and CIFAR-$100$ datasets, where $T$ is fixed to $1000$, $\xi$ is set to $0.0014$, and $\tau$ is chosen such that $10\%$ of the activation values are clamped, following the setups of the original papers. The performance improvements are illustrious regarding the performance of MaxLogit, ReAct, and ODIN with and without watermarking, largely confirming the fact that our proposed watermarking is orthogonal to existing works. {Further, since we directly use the training strategy of softmax scoring-based watermarking for MaxLogit, our results in Table~\ref{tab: maxlogit} demonstrate that watermarking can also benefit from improved choices of scoring strategies. }

\begin{table}[t]
\parbox{.47\linewidth}{
\centering
\caption{ReAct scoring with/without watermarking on CIFAR benchmarks. $\downarrow$ ($\uparrow$) indicates smaller (larger) values are preferred.} \label{tab: react}
\vspace{5pt}
\scriptsize
{
\begin{tabular}{c|ccc}
\toprule[1.5pt]
                   & FPR95 $\downarrow$     & AUROC $\uparrow$       & AUPR $\uparrow$      \\
                   \cline{2-4} 
\multirow{-2}{*}{} & \multicolumn{3}{c}{w/ (w/o) watermark} \\
\midrule[0.6pt]
\multicolumn{4}{c}{\cellcolor{greyL}CIFAR-10} \\
\midrule[0.6pt]
iSUN               & \textbf{27.90} (63.65) & \textbf{95.73} (87.54) & \textbf{99.13} (97.17) \\ 
Places$365$        & \textbf{62.55} (62.65) & \textbf{85.90} (86.98) & \textbf{96.50} (96.74) \\ 
Texture            & \textbf{39.85} (58.90) & \textbf{93.68} (87.32) & \textbf{98.63} (96.86) \\ 
SVHN               & \textbf{40.40} (43.35) & \textbf{93.47} (93.27) & \textbf{98.61} (98.60) \\ 
LSUN             & \textbf{23.35} (59.40) & \textbf{96.29} (88.85) & \textbf{9.24} (97.51) \\ 
\midrule
\textbf{average}   & \textbf{38.91} (57.59) & \textbf{93.01} (88.79) & \textbf{80.42} (97.37) \\  \midrule[1pt]
\multicolumn{4}{c}{\cellcolor{greyL}CIFAR-100} \\
\midrule[1pt]
iSUN               & \textbf{68.05} (86.40) & \textbf{83.91} (75.31) & \textbf{96.35} (94.04) \\ 
Places$365$        & \textbf{82.65} (87.70) & \textbf{73.18} (71.20) & \textbf{93.00} (92.50) \\ 
Texture            & \textbf{74.65} (86.35) & \textbf{78.35} (71.33) & \textbf{94.72} (92.39) \\ 
SVHN               & \textbf{85.95} (77.50) & \textbf{75.47} (72.79) & \textbf{94.34} (92.33) \\ 
LSUN               & \textbf{66.95} (86.85) & \textbf{84.02} (74.71) & \textbf{96.39} (93.90) \\ 
\midrule
\textbf{average}   & \textbf{75.65} (84.96) & \textbf{78.98} (73.06) & \textbf{94.96} (93.03) \\  \bottomrule[1.5pt]      
\end{tabular}
}}~~~~
\parbox{.47\linewidth}{
\centering
\caption{ODIN scoring with/without watermarking on CIFAR benchmarks. $\downarrow$ ($\uparrow$) indicates smaller (larger) values are preferred.} \label{tab: odin}
\vspace{5pt}
\scriptsize{
\begin{tabular}{c|ccc}
\toprule[1.5pt]
                   & FPR$95$ $\downarrow$     & AUROC $\uparrow$       & AUPR $\uparrow$      \\
                   \cline{2-4} 
\multirow{-2}{*}{} & \multicolumn{3}{c}{w/ (w/o) watermark} \\
\midrule[0.6pt]
\multicolumn{4}{c}{\cellcolor{greyL}CIFAR-10} \\
\midrule[0.6pt]
iSUN               & \textbf{25.05} (35.15) & \textbf{96.21} (93.09) & \textbf{99.23} (98.44) \\ 
Places$365$        & \textbf{59.60} (55.95) & \textbf{87.74} (85.84) & \textbf{97.07} (96.22) \\ 
Texture            & \textbf{36.35} (49.50) & \textbf{93.83} (86.72) & \textbf{98.64} (96.18) \\ 
SVHN               & \textbf{40.55} (43.20) & \textbf{93.29} (91.34) & \textbf{98.56} (97.95) \\ 
LSUN             & \textbf{23.75} (29.40) & \textbf{96.45} (94.06) & \textbf{99.27} (98.64) \\ 
\midrule
\textbf{average}   & \textbf{37.06} (42.64) & \textbf{93.50} (90.21) & \textbf{98.55} (97.49) \\  \midrule[1pt]
\multicolumn{4}{c}{\cellcolor{greyL}CIFAR-100} \\
\midrule[1pt]
iSUN               & \textbf{69.60} (70.80) & \textbf{83.70} (81.32) & \textbf{96.32} (95.39) \\ 
Places$365$        & \textbf{82.10} (88.50) & \textbf{74.65} (72.07) & \textbf{93.64} (92.74) \\ 
Texture            & \textbf{75.95} (82.40) & \textbf{78.80} (71.87) & \textbf{94.86} (92.50) \\ 
SVHN               & \textbf{86.35} (94.65) & \textbf{76.04} (59.40) & \textbf{94.52} (89.21) \\ 
LSUN               & \textbf{67.80} (71.20) & \textbf{84.61} (81.18) & \textbf{96.60} (95.47) \\ 
\midrule
\textbf{average}   & \textbf{76.36} (81.51) & \textbf{79.56} (73.17) & \textbf{95.18} (93.06) \\  \bottomrule[1.5pt]      
\end{tabular}
}}
\end{table}

\subsection{Other Learning Strategies with Watermarking}

Also, we consider the situation where a set of OOD data are available for watermark training, where we adopt the tiny-ImageNet~\cite{le2015tiny} dataset as training-time OOD dataset. Here, we replace the Gaussian noise in the OOD learning objective $\ell^\text{OOD}$ to be the randomly selected sample from the tiny-ImageNet dataset. Then, the learning objective with training-time OOD data is of the form
\begin{align}
    \mathcal{L}_n(\boldsymbol{w})=\sum_n \ell^\text{ID}(\boldsymbol{x}_i + \boldsymbol{w},y_i;f) + \beta \sum_n \ell^\text{OOD}(\boldsymbol{o}_j+ \boldsymbol{w};f), \label{eq: objective_oe}
\end{align}
where $\boldsymbol{o}_j$ denotes the randomly selected sample from the tiny-ImageNet dataset. The experimental results on CIFAR-$10$ and CIFAR-$100$ datasets are summarized in Table~\ref{tab: softmax tiny} and Table~\ref{tab: free energy tiny}. Unfortunately, for our current realization, we observe that watermark training with extra OOD data fails to induce a large performance improvement in OOD detection. Even worse, in some cases, learning with extra OOD data can impair the power of the resultant watermarks. We conjecture that the inductive bias introduced by the training-time OOD data may deviate from the test-time data, severely misleading the resultant watermarks in showing results even lower than that of the simple Gaussian noise.  

\begin{table}[t]
\parbox{.47\linewidth}{
\centering
\caption{Softmax scoring with tiny-ImageNet and Gaussian noise on CIFAR benchmarks. $\downarrow$ ($\uparrow$) indicates smaller (larger) values are preferred.} \label{tab: softmax tiny}
\vspace{5pt}
\scriptsize
{
\begin{tabular}{c|ccc}
\toprule[1.5pt]
                   & FPR95 $\downarrow$     & AUROC $\uparrow$       & AUPR $\uparrow$      \\
                   \cline{2-4} 
\multirow{-2}{*}{} & \multicolumn{3}{c}{tiny-ImageNet (Gaussian Noise)} \\
\midrule[0.6pt]
\multicolumn{4}{c}{\cellcolor{greyL}CIFAR-10} \\
\midrule[0.6pt]
iSUN               & \textbf{22.80} (43.60) & \textbf{95.88} (93.53) & \textbf{99.16} (98.67) \\ 
Places$365$        & \textbf{61.95} (60.75) & {86.47} (\textbf{87.85}) & {96.87} (\textbf{96.94}) \\
Texture            & \textbf{41.45} (42.00) & \textbf{92.93} (92.83) & \textbf{98.51} (98.43) \\
SVHN               & {67.15} (\textbf{27.25}) & {88.43} (\textbf{96.00}) & {97.59} (\textbf{99.17}) \\
LSUN             & \textbf{22.40} (40.70) & \textbf{96.16} (94.36) & \textbf{99.22} (98.86) \\
\midrule
\textbf{average}   & {43.15} (\textbf{42.86}) & {91.97} (\textbf{92.91}) & {98.27} (\textbf{98.41}) \\ \midrule[1pt]
\multicolumn{4}{c}{\cellcolor{greyL}CIFAR-100} \\
\midrule[1pt]
iSUN               & \textbf{72.85} (77.85) & \textbf{81.22} (79.91) & \textbf{95.66} (95.35) \\ 
Places$365$        & \textbf{79.75} (83.25) & \textbf{75.55} (74.53) & \textbf{93.77} (93.47) \\
Texture            & \textbf{73.35} (78.10) & \textbf{78.88} (77.14) & \textbf{94.61} (94.26) \\
SVHN               & {83.00} (\textbf{82.95}) & {76.15} (\textbf{76.92}) & {94.52} (\textbf{94.72}) \\
LSUN               & \textbf{71.65} (76.75) & \textbf{82.93} (79.60) & \textbf{96.16} (95.27) \\
\midrule
\textbf{average}   & \textbf{76.20} (76.54) & \textbf{79.10} (78.99) & \textbf{95.08} (94.96) \\ \bottomrule[1.5pt]      
\end{tabular}
}}~~~~
\parbox{.47\linewidth}{
\centering
\caption{Free energy scoring with tiny-ImageNet and Gaussian noise on CIFAR benchmarks. $\downarrow$ ($\uparrow$) indicates smaller (larger) values are preferred.} \label{tab: free energy tiny}
\vspace{5pt}
\scriptsize{
\begin{tabular}{c|ccc}
\toprule[1.5pt]
                   & FPR$95$ $\downarrow$     & AUROC $\uparrow$       & AUPR $\uparrow$      \\
                   \cline{2-4} 
\multirow{-2}{*}{} & \multicolumn{3}{c}{tiny-ImageNet (Gaussian Noise)} \\
\midrule[0.6pt]
\multicolumn{4}{c}{\cellcolor{greyL}CIFAR-10} \\
\midrule[0.6pt]
iSUN               & {23.65} (\textbf{16.30}) & {95.89} (\textbf{96.97}) & {99.16} (\textbf{99.39}) \\
Places$365$        & \textbf{36.15} (36.25) & \textbf{91.89} (91.87) & \textbf{97.94} (97.94) \\
Texture            & \textbf{31.20} (32.60) & \textbf{93.34} (93.14) & \textbf{98.39} (98.08) \\
SVHN               & \textbf{13.80} (16.45) & \textbf{97.53} (97.11) & \textbf{99.49} (99.39) \\
LSUN             & {23.15} (\textbf{16.85}) & {86.20} (\textbf{96.97}) & {99.20} (\textbf{99.38}) \\
\midrule
\textbf{average}   & {25.59} (\textbf{23.69}) & {92.97} (\textbf{95.21}) & \textbf{98.83} (98.83) \\ \midrule[1pt]
\multicolumn{4}{c}{\cellcolor{greyL}CIFAR-100} \\
\midrule[1pt]
iSUN               & \textbf{77.10} (75.05) & \textbf{83.62} (83.07) & \textbf{96.40} (96.15) \\ 
Places$365$        & \textbf{79.25} (80.50) & {77.67} (\textbf{77.78}) & {94.22} (\textbf{94.45}) \\
Texture            & \textbf{68.70} (75.15) & \textbf{81.36} (79.55) & \textbf{95.10} (94.79) \\
SVHN               & \textbf{82.00} (82.85) & \textbf{77.38} (75.26) & \textbf{94.80} (94.18) \\
LSUN               & \textbf{71.35} (71.85) & \textbf{84.14} (84.01) & \textbf{96.55} (96.33) \\
\midrule
\textbf{average}   & \textbf{75.68} (77.08) & \textbf{80.83} (79.93) & \textbf{95.41} (95.18) \\ \bottomrule[1.5pt]      
\end{tabular}
}}
\end{table}

{We also list the detection performance with ``perm'' and ``rotate'' in Section~\ref{sec: experiment}, demonstrating the effectiveness of the resultant watermarks on far OOD cases. Here, the training objective is:}
\begin{align}
    {\mathcal{L}_n(\boldsymbol{w})=\sum_n \ell^\text{ID}(\boldsymbol{x}_i + \boldsymbol{w},y_i;f) + \beta \sum_n \ell^\text{OOD}(\boldsymbol{\epsilon}_j+ \boldsymbol{w};f) + \beta \sum_n \ell^\text{OOD}(\boldsymbol{\tilde{x}}_i+ \boldsymbol{w};f),} \label{eq: objective v2}
\end{align}
{where $\boldsymbol{\tilde{x}}$ is an augmentation of the original $\boldsymbol{x}$ with either ``perm'' or ``rotate''. The results are summarized from Tables~\ref{tab: softmax perm} to~\ref{tab: eneregy rotate}. As we can see, the results with "perm" and "rotate" is comparable with (even better than) the original learning setup with only the Gaussian noise (common). }

\begin{table}[t]
\parbox{.47\linewidth}{
\centering
\caption{{Softmax scoring with learning from Gaussian noise and ``perm'' augmentation. $\downarrow$ ($\uparrow$) indicates smaller (larger) values are preferred.}} \label{tab: softmax perm}
\vspace{5pt}
\scriptsize
{
\begin{tabular}{c|ccc}
\toprule[1.5pt]
                   & FPR95 $\downarrow$     & AUROC $\uparrow$       & AUPR $\uparrow$      \\
                   \cline{2-4} 
\multirow{-2}{*}{} & \multicolumn{3}{c}{perm (common)} \\
\midrule[0.6pt]
\multicolumn{4}{c}{\cellcolor{greyL}CIFAR-10} \\
\midrule[0.6pt]
iSUN               & \textbf{38.00} (41.50) & \textbf{93.99} (93.98) & \textbf{98.79} (98.77) \\ 
Places$365$        & \textbf{55.20} (56.30) & \textbf{89.13} (89.03) & \textbf{97.49} (97.32) \\
Texture            & \textbf{42.20} (43.80) & 92.30 (\textbf{93.06}) & 98.28 (\textbf{98.46}) \\
SVHN               & 33.35 (\textbf{27.00}) & 94.75 (\textbf{96.07}) & {98.93} (\textbf{99.20}) \\
LSUN               & \textbf{36.40} (37.85) & 94.17 (\textbf{94.57}) & 98.80 (\textbf{98.89}) \\
\midrule
\textbf{average}   & \textbf{41.03} (41.29) & {92.87} (\textbf{93.34}) & {98.46} (\textbf{98.53}) \\ \midrule[1pt]
\multicolumn{4}{c}{\cellcolor{greyL}CIFAR-100} \\
\midrule[1pt]
iSUN               & \textbf{77.90} (84.00) & \textbf{79.11} (75.98) & \textbf{95.11} (94.37) \\ 
Places$365$        & \textbf{79.90} (83.60) & \textbf{75.21} (73.20) & \textbf{93.68} (93.22) \\
Texture            & \textbf{75.80} (83.00) & \textbf{77.13} (72.45) & \textbf{94.17} (92.78) \\
SVHN               & \textbf{85.45} (87.20) & \textbf{73.22} (72.45) & \textbf{93.65} (93.45) \\
LSUN               & \textbf{76.85} (81.05) & \textbf{79.37} (77.40) & \textbf{95.21} (94.75) \\
\midrule
\textbf{average}   & \textbf{79.18} (83.77) & \textbf{76.81} (74.30) & \textbf{94.36} (93.71) \\ \bottomrule[1.5pt]      
\end{tabular}
}}~~~~
\parbox{.47\linewidth}{
\centering
\caption{{Softmax scoring with learning from Gaussian noise and ``rotate'' augmentation. $\downarrow$ ($\uparrow$) indicates smaller (larger) values are preferred.}} \label{tab: softmax rotate}
\vspace{5pt}
\scriptsize{
\begin{tabular}{c|ccc}
\toprule[1.5pt]
                   & FPR95 $\downarrow$     & AUROC $\uparrow$       & AUPR $\uparrow$      \\
                   \cline{2-4} 
\multirow{-2}{*}{} & \multicolumn{3}{c}{rotate (common)} \\
\midrule[0.6pt]
\multicolumn{4}{c}{\cellcolor{greyL}CIFAR-10} \\
\midrule[0.6pt]
iSUN               & \textbf{40.25} (41.50) & 93.44 (\textbf{93.98}) & 98.63 (\textbf{98.77}) \\ 
Places$365$        & \textbf{56.15} (56.30) & 88.12 (\textbf{89.03}) & 97.11 (\textbf{97.32}) \\
Texture            & \textbf{41.15} (43.80) & 92.66 (\textbf{93.06}) & 98.34 (\textbf{98.46}) \\
SVHN               & 29.65 (\textbf{27.00}) & 95.42 (\textbf{96.07}) & {99.07} (\textbf{99.20}) \\
LSUN               & 37.90 (\textbf{37.85}) & 93.95 (\textbf{94.57}) & 98.76 (\textbf{98.89}) \\
\midrule
\textbf{average}   & \textbf{41.02} (41.29) & {92.72} (\textbf{93.34}) & {98.38} (\textbf{98.53}) \\ \midrule[1pt]
\multicolumn{4}{c}{\cellcolor{greyL}CIFAR-100} \\
\midrule[1pt]
iSUN               & \textbf{77.45} (84.00) & \textbf{78.73} (75.98) & \textbf{94.95} (94.37) \\ 
Places$365$        & \textbf{79.35} (83.60) & \textbf{75.26} (73.20) & \textbf{93.67} (93.22) \\
Texture            & \textbf{76.35} (83.00) & \textbf{76.88} (72.45) & \textbf{94.16} (92.78) \\
SVHN               & \textbf{84.90} (87.20) & \textbf{72.98} (72.45) & \textbf{93.58} (93.45) \\
LSUN               & \textbf{78.10} (81.05) & \textbf{78.59} (77.40) & \textbf{95.00} (94.75) \\
\midrule
\textbf{average}   & \textbf{79.23} (83.77) & \textbf{76.49} (74.30) & \textbf{94.27} (93.71) \\ \bottomrule[1.5pt]      
\end{tabular}
}}
\end{table}

\begin{table}[t]
\parbox{.47\linewidth}{
\centering
\caption{{Free energy scoring with learning from Gaussian noise and ``perm'' augmentation. $\downarrow$ ($\uparrow$) prefers smaller (larger) values.}} \label{tab: eneregy perm}
\vspace{5pt}
\scriptsize
{
\begin{tabular}{c|ccc}
\toprule[1.5pt]
                   & FPR95 $\downarrow$     & AUROC $\uparrow$       & AUPR $\uparrow$      \\
                   \cline{2-4} 
\multirow{-2}{*}{} & \multicolumn{3}{c}{perm (common)} \\
\midrule[0.6pt]
\multicolumn{4}{c}{\cellcolor{greyL}CIFAR-10} \\
\midrule[0.6pt]
iSUN               & 25.60 (\textbf{24.20}) & 95.81 (\textbf{96.10}) & 99.14 (\textbf{99.22}) \\ 
Places$365$        & 39.70 (\textbf{39.45}) & \textbf{91.96} (91.46) & \textbf{98.01} (97.89) \\
Texture            & 39.15 (\textbf{38.95}) & 92.13 (\textbf{92.65}) & 97.93 (\textbf{98.16}) \\
SVHN               & \textbf{16.95} (18.75) & \textbf{97.01} (96.54) & \textbf{99.37} (99.27) \\
LSUN               & 22.10 (\textbf{21.80}) & \textbf{96.38} (96.27) & \textbf{99.27} (99.24) \\
\midrule
\textbf{average}   & 28.70 (\textbf{28.63}) & \textbf{94.66} (94.61) & {98.74} (\textbf{98.76}) \\ \midrule[1pt]
\multicolumn{4}{c}{\cellcolor{greyL}CIFAR-100} \\
\midrule[1pt]
iSUN               & 77.05 (\textbf{75.30}) & 83.49 (\textbf{84.51}) & 96.32 (\textbf{96.78}) \\ 
Places$365$        & 80.40 (\textbf{78.05}) & 77.11 (\textbf{78.15}) & 94.25 (\textbf{94.28}) \\
Texture            & \textbf{68.85} (70.80) & \textbf{81.31} (81.14) & 95.14 (\textbf{95.02}) \\
SVHN               & 81.95 (\textbf{80.50}) & \textbf{78.27} (77.27) & \textbf{95.14} (94.70) \\
LSUN               & 77.20 (\textbf{75.10}) & \textbf{83.58} (83.53) & 96.35 (\textbf{97.05}) \\
\midrule
\textbf{average}   & 77.09 (\textbf{75.95}) & 80.75 (\textbf{81.12}) & 95.44 (\textbf{95.57}) \\ \bottomrule[1.5pt]      
\end{tabular}
}}~~~~
\parbox{.47\linewidth}{
\centering
\caption{{Free energy scoring with learning from Gaussian noise and ``rotate'' augmentation. $\downarrow$ ($\uparrow$) prefers smaller (larger) values.}} \label{tab: eneregy rotate}
\vspace{5pt}
\scriptsize{
\begin{tabular}{c|ccc}
\toprule[1.5pt]
                   & FPR95 $\downarrow$     & AUROC $\uparrow$       & AUPR $\uparrow$      \\
                   \cline{2-4} 
\multirow{-2}{*}{} & \multicolumn{3}{c}{rotate (common)} \\
\midrule[0.6pt]
\multicolumn{4}{c}{\cellcolor{greyL}CIFAR-10} \\
\midrule[0.6pt]
iSUN               & 23.85 (\textbf{24.20}) & 96.04 (\textbf{96.10}) & 99.20 (\textbf{99.22}) \\ 
Places$365$        & \textbf{38.60} (39.45) & \textbf{91.94} (91.46) & \textbf{97.97} (97.89) \\
Texture            & \textbf{35.20} (38.95) & \textbf{92.93} (92.65) & \textbf{98.26} (98.16) \\
SVHN               & \textbf{16.85} (18.75) & \textbf{97.00} (96.54) & \textbf{99.37} (99.27) \\
LSUN               & 22.80 (\textbf{21.80}) & 96.10 (\textbf{96.27}) & 99.19 (\textbf{99.24}) \\
\midrule
\textbf{average}   & \textbf{27.46} (28.63) & \textbf{94.80} (94.61) & {98.80} (\textbf{98.76}) \\ \midrule[1pt]
\multicolumn{4}{c}{\cellcolor{greyL}CIFAR-100} \\
\midrule[1pt]
iSUN               & 85.45 (\textbf{75.30}) & 81.39 (\textbf{84.51}) & 95.85 (\textbf{96.78}) \\ 
Places$365$        & 80.10 (\textbf{78.05}) & 76.53 (\textbf{78.15}) & 93.79 (\textbf{94.28}) \\
Texture            & 71.55 (\textbf{70.80}) & 80.27 (\textbf{81.14}) & 94.67 (\textbf{95.02}) \\
SVHN               & \textbf{80.20} (80.50) & \textbf{79.27} (77.27) & \textbf{95.29} (94.70) \\
LSUN               & 81.25 (\textbf{75.10}) & 81.85 (\textbf{83.53}) & 95.97 (\textbf{97.05}) \\
\midrule
\textbf{average}   & 79.71 (\textbf{75.95}) & 79.86 (\textbf{81.12}) & 95.12 (\textbf{95.57}) \\ \bottomrule[1.5pt]      
\end{tabular}
}}
\end{table}

\subsection{Comparison with State-of-the-art Methods}

For the concreteness of our discussion, this section compares our proposal with state-of-the-art methods in OOD detection. In particular, we compare with softmax scoring (Softmax)~\cite{hendrycks2016baseline}, free energy scoring (Energy)~\cite{liu2020energy}, ReAct~\cite{sun2021react}, ODIN~\cite{LiangLS18}, Mahalanobis~\cite{lee2018simple}, GradNorm~\cite{huang2021importance}, and OE~\cite{HendrycksMD19}. The experimental results on CIFAR benchmarks are summarized in Table~\ref{tab: sota}. The average performance on iSUN, Places$365$, Texture, SVHN, and LSUN is reported. As we can see, watermarking can boost the performance of various scoring methods in OOD detection, achieving the best detection performance compared with all other advanced methods.

\begin{table}[t]
\centering
\caption{Comparison of watermarking and different OOD scoring functions on CIFAR benchmarks.} \label{tab: sota}
\resizebox{\columnwidth}{!}{
\scriptsize{
\begin{tabular}{c|ccccccc|ccccc}
\toprule[1.5pt]
        & \multicolumn{7}{c|}{w/o watermarking}                          & \multicolumn{5}{c}{w/ watermarking}  \\
        \cline{2-13}
        & \makecell{Softmax\\\cite{hendrycks2016baseline}} & \makecell{Energy\\\cite{liu2020energy}} & \makecell{ReAct\\\cite{sun2021react}} & \makecell{ODIN\\\cite{LiangLS18}} &  \makecell{Mahalan-\\obis~\cite{lee2018simple}} & \makecell{GradNorm\\\cite{cimpoi2014describing}} & \makecell{OE\\\cite{HendrycksMD19}} & Softmax & Energy & ReAct & ODIN & OE  \\
\midrule
\multicolumn{13}{c}{\cellcolor{greyL}CIFAR-$10$}  \\
\midrule
FPR$95$ & 55.70 & 37.67 & 57.59 & 42.64 & 34.18 & 40.51 & 20.96 & 42.86 & 23.69 & 38.91 & 37.06 & \textbf{19.06} \\
AUROC   & 89.82 & 90.56 & 88.79 & 90.21 & 93.23 & 90.10 & 96.41 & 92.91 & 95.21 & 93.01 & 93.50 & \textbf{96.89} \\
AUPR    & 97.32 & 97.46 & 97.37 & 97.49 & 98.41 & 97.35 & 99.39 & 98.42 & 98.83 & 80.42 & 98.55 & \textbf{99.31} \\
\bottomrule[1pt] 
\multicolumn{13}{c}{\cellcolor{greyL}CIFAR-$100$}   \\
\midrule
FPR$95$ & 82.97 & 81.61 & 84.96 & 81.51 & 55.63 & 83.68 & 60.30 & 79.98 & 77.07 & 75.65 & 76.36 & \textbf{54.20} \\
AUROC   & 73.88 & 76.50 & 73.06 & 73.17 & 82.26 & 72.93 & 82.99 & 77.57 & 79.93 & 78.98 & 79.56 & \textbf{83.90} \\
AUPR    & 93.43 & 94.05 & 93.03 & 93.06 & 95.56 & 93.00 & 95.91 & 94.61 & 95.18 & 94.96 & 95.18 & \textbf{96.08} \\
\bottomrule[1.5pt]      
\end{tabular}}}
\end{table}

\subsection{Experiments with Mean and Standard Deviation}

\begin{figure*}[!htp]
\centering  
\subfigure[softmax scoring on CIFAR-$10$]{
\centering  
\begin{minipage}[t]{\linewidth}
    \centering
	\includegraphics[width=.9\linewidth]{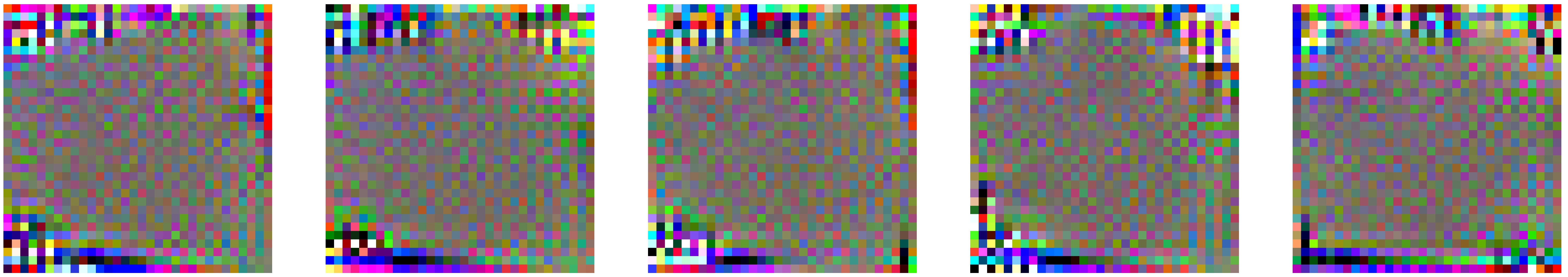}
    \centering  
\end{minipage}} 

\subfigure[softmax scoring on CIFAR-$100$]{
\centering  
\begin{minipage}[t]{\linewidth}
    \centering  
    \includegraphics[width=.9\linewidth]{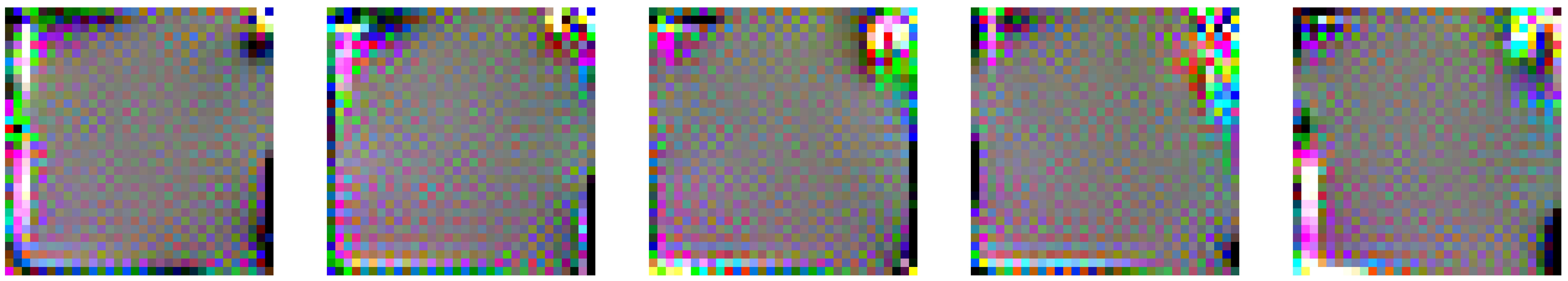}
    \centering
\end{minipage}}

\subfigure[free energy scoring on CIFAR-$10$]{
\centering  
\begin{minipage}[t]{\linewidth}
    \centering  
    \includegraphics[width=.9\linewidth]{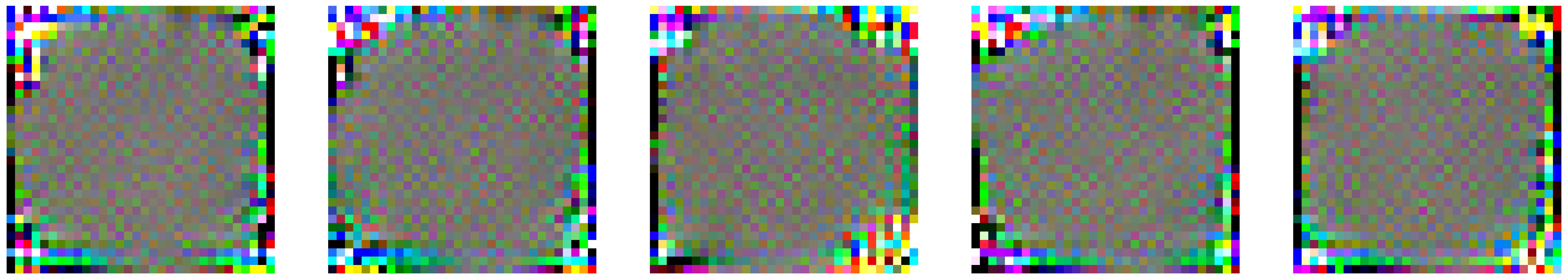}
    \centering
\end{minipage}}

\subfigure[free energy scoring on CIFAR-$100$]{
\centering  
\begin{minipage}[t]{\linewidth}
    \centering  
    \includegraphics[width=.9\linewidth]{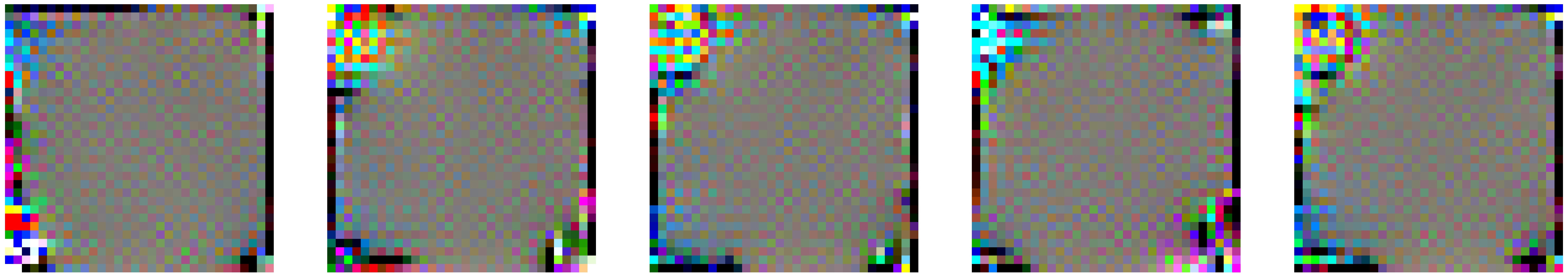}
    \centering
\end{minipage}}

\caption{Illustrations of the learned watermarks with $5$ individual trails.} 
\label{fig: f}
\end{figure*}

This section further verifies the results from Table~\ref{tab: softmax} to Table~\ref{tab: imagenet_fe} with five individual trails (random seeds). In Table~\ref{tab: softmax_ms}, Table~\ref{tab: free energy_ms}, and Table~\ref{tab: imagenet}, we summarize the average results and the standard deviation for the softmax scoring and the free energy scoring, respectively. In Figure~\ref{fig: f}, we also depict the learned watermarks for each trial on CIFAR-$10$ and CIFAR-$100$.  Overall, we observe that the learned watermarks preserve some similar pattern (e.g., the shape of areas with large values) given the same dataset and the same scoring strategy, and the improvement of watermarking is stable across different datasets and scoring methods. 

\begin{table}[t]
\centering
\caption{The softmax scoring with/without watermarking on CIFAR benchmarks. Five individual trails (mean $\pm$ std) are conducted. The notion $\downarrow$ ($\uparrow$) indicates smaller (larger) values are preferred.} \label{tab: softmax_ms}
\vspace{5pt}
\scriptsize
{
\begin{tabular}{c|ccc}
\toprule[1.5pt]
                   & FPR95 $\downarrow$     & AUROC $\uparrow$       & AUPR $\uparrow$      \\
                   \cline{2-4} 
\multirow{-2}{*}{} & \multicolumn{3}{c}{w/ (w/o) watermark} \\
\midrule[0.6pt]
\multicolumn{4}{c}{\cellcolor{greyL}CIFAR-10} \\
\midrule[0.6pt]
iSUN               & \textbf{44.68 $\pm$ 1.49} (55.43 $\pm$ 0.29) & \textbf{93.38 $\pm$ 0.28} (90.10 $\pm$ 0.22) & \textbf{97.80 $\pm$ 0.07} (97.80 $\pm$ 0.07) \\ 
Places$365$        & \textbf{59.21 $\pm$ 0.97} (60.53 $\pm$ 1.31) & \textbf{88.83 $\pm$ 0.34} (87.83 $\pm$ 0.17) & \textbf{97.08 $\pm$ 0.11} (97.06 $\pm$ 0.06) \\
Texture            & \textbf{42.07 $\pm$ 1.23} (59.37 $\pm$ 1.55) & \textbf{93.03 $\pm$ 0.23} (88.56 $\pm$ 0.37) & \textbf{98.47 $\pm$ 0.07} (97.20 $\pm$ 0.09) \\
SVHN               & \textbf{29.25 $\pm$ 2.17} (48.07 $\pm$ 0.97) & \textbf{95.69 $\pm$ 0.35} (91.80 $\pm$ 0.13) & \textbf{99.11 $\pm$ 0.07} (97.24 $\pm$ 0.03) \\
LSUN             & \textbf{40.45 $\pm$ 2.14} (52.23 $\pm$ 1.04) & \textbf{94.11 $\pm$ 0.45} (91.50 $\pm$ 0.10) & \textbf{98.79 $\pm$ 0.11} (98.15 $\pm$ 0.04) \\
\midrule
\textbf{average}   & \textbf{43.13 $\pm$ 0.16} (55.12 $\pm$ 0.10) & \textbf{93.00 $\pm$ 0.33} (89.95 $\pm$ 0.19) & \textbf{98.25 $\pm$ 0.08} (97.49 $\pm$ 0.05) \\ \midrule[1pt]
\multicolumn{4}{c}{\cellcolor{greyL}CIFAR-100} \\
\midrule[1pt]
iSUN               & \textbf{78.70 $\pm$ 1.48} (82.40 $\pm$ 0.81) & \textbf{78.30 $\pm$ 0.57} (75.62 $\pm$ 0.33) & \textbf{94.92 $\pm$ 0.17} (94.11 $\pm$ 0.10) \\ 
Places$365$        & \textbf{82.55 $\pm$ 0.65} (82.97 $\pm$ 0.86) & \textbf{74.69 $\pm$ 0.50} (74.29 $\pm$ 0.29) & \textbf{93.77 $\pm$ 0.13} (93.41 $\pm$ 0.12) \\
Texture            & \textbf{77.83 $\pm$ 1.47} (83.48 $\pm$ 0.70) & \textbf{76.42 $\pm$ 0.18} (73.37 $\pm$ 0.37) & \textbf{94.06 $\pm$ 0.08} (92.95 $\pm$ 0.11) \\
SVHN               & \textbf{83.71 $\pm$ 2.27} (84.72 $\pm$ 0.73) & \textbf{76.16 $\pm$ 0.78} (71.29 $\pm$ 0.63) & \textbf{94.51 $\pm$ 0.17} (92.88 $\pm$ 0.22) \\
LSUN             & \textbf{78.57 $\pm$ 1.09} (81.67 $\pm$ 0.78) & \textbf{78.37 $\pm$ 0.54} (75.77 $\pm$ 0.33) & \textbf{94.92 $\pm$ 0.16} (94.18 $\pm$ 0.11) \\
\midrule
\textbf{average}   & \textbf{88.27 $\pm$ 1.39} (83.04 $\pm$ 0.77) & \textbf{76.78 $\pm$ 0.51} (74.06 $\pm$ 0.39) & \textbf{94.43 $\pm$ 0.14} (93.50 $\pm$ 0.13) \\ \bottomrule[1.5pt]      
\end{tabular}
}
\end{table}

\begin{table}[!thp]
\centering
\caption{The free energy scoring with/without watermarking on CIFAR benchmarks. Five individual trails (mean $\pm$ std) are conducted. The notion $\downarrow$ ($\uparrow$) indicates smaller (larger) values are preferred.} \label{tab: free energy_ms}
\vspace{5pt}
\scriptsize{
\begin{tabular}{c|ccc}
\toprule[1.5pt]
                   & FPR$95$ $\downarrow$     & AUROC $\uparrow$       & AUPR $\uparrow$      \\
                   \cline{2-4} 
\multirow{-2}{*}{} & \multicolumn{3}{c}{w/ (w/o) watermark} \\
\midrule[0.6pt]
\multicolumn{4}{c}{\cellcolor{greyL}CIFAR-10} \\
\midrule[0.6pt]
iSUN               & \textbf{18.84 $\pm$ 2.76} (33.66 $\pm$ 0.75) & \textbf{96.28 $\pm$ 0.62} (92.62 $\pm$ 0.31) & \textbf{99.23 $\pm$ 0.13} (98.27 $\pm$ 0.10) \\
Places$365$          & \textbf{38.89 $\pm$ 1.74} (40.67 $\pm$ 0.91) & \textbf{91.92 $\pm$ 0.41} (89.62 $\pm$ 0.20) & \textbf{98.01 $\pm$ 0.11} (97.16 $\pm$ 0.12) \\
Texture            & \textbf{34.60 $\pm$ 2.16} (52.67 $\pm$ 1.10) & \textbf{93.36 $\pm$ 0.26} (85.19 $\pm$ 0.35) & \textbf{98.31 $\pm$ 0.06} (95.40 $\pm$ 0.13) \\
SVHN               & \textbf{14.96 $\pm$ 0.93} (35.60 $\pm$ 0.50) & \textbf{97.12 $\pm$ 0.15} (91.08 $\pm$ 0.22) & \textbf{99.39 $\pm$ 0.03} (97.71 $\pm$ 0.07) \\
LSUN             & \textbf{16.63 $\pm$ 2.12} (27.12 $\pm$ 0.85) & \textbf{96.43 $\pm$ 0.43} (94.32 $\pm$ 0.07) & \textbf{99.26 $\pm$ 0.09} (98.70 $\pm$ 0.02) \\
\midrule
\textbf{average}   & \textbf{24.78 $\pm$ 1.94} (37.94 $\pm$ 0.82) & \textbf{95.02 $\pm$ 0.37} (90.56 $\pm$ 0.23) & \textbf{98.84 $\pm$ 0.08} (97.44 $\pm$ 0.08) \\ \midrule[1pt]
\multicolumn{4}{c}{\cellcolor{greyL}CIFAR-100} \\
\midrule[1pt]
iSUN               & \textbf{74.62 $\pm$ 1.97} (81.85 $\pm$ 1.14) & \textbf{84.30 $\pm$ 0.85} (78.78 $\pm$ 0.40) & \textbf{96.53 $\pm$ 0.24} (94.90 $\pm$ 0.13) \\ 
Places$365$        & \textbf{77.79 $\pm$ 0.27} (80.27 $\pm$ 1.02) & \textbf{78.13 $\pm$ 0.78} (76.46 $\pm$ 0.54) & \textbf{94.40 $\pm$ 0.31} (93.92 $\pm$ 0.16) \\
Texture            & \textbf{68.96 $\pm$ 1.51} (79.47 $\pm$ 0.27) & \textbf{82.07 $\pm$ 0.62} (76.34 $\pm$ 0.34) & \textbf{95.38 $\pm$ 0.24} (93.64 $\pm$ 0.11) \\
SVHN               & \textbf{80.30 $\pm$ 0.75} (85.80 $\pm$ 0.87) & \textbf{78.55 $\pm$ 0.63} (73.61 $\pm$ 0.37) & \textbf{95.11 $\pm$ 0.18} (93.51 $\pm$ 0.10) \\
LSUN             & \textbf{71.25 $\pm$ 2.00} (79.26 $\pm$ 1.43) & \textbf{84.94 $\pm$ 0.54} (79.34 $\pm$ 0.40) & \textbf{96.65 $\pm$ 0.12} (94.99 $\pm$ 0.11) \\
\midrule
\textbf{average}   & \textbf{74.58 $\pm$ 1.30} (81.33 $\pm$ 0.94) & \textbf{81.59 $\pm$ 0.68} (76.90 $\pm$ 0.41) & \textbf{95.61 $\pm$ 0.21} (94.19 $\pm$ 0.12) \\ \bottomrule[1.5pt]      
\end{tabular}
}
\end{table}




\begin{table}[t]
\centering
\caption{Softmax and free energy scoring with/without watermarking on ImageNet. Five individual trails (mean $\pm$ std) are conducted. The notion $\downarrow$ ($\uparrow$) indicates smaller (larger) values are preferred.} \label{tab: imagenet}
\vspace{5pt}
\scriptsize{
\begin{tabular}{c|ccc}
\toprule[1.5pt]
                   & FPR$95$ $\downarrow$     & AUROC $\uparrow$       & AUPR $\uparrow$      \\
                   \cline{2-4} 
\multirow{-2}{*}{} & \multicolumn{3}{c}{w/ (w/o) watermark} \\
\midrule[0.6pt]
\multicolumn{4}{c}{\cellcolor{greyL}Softmax Scoring} \\
\midrule[0.6pt]
iSUN               & \textbf{12.69 $\pm$ 1.55} (52.38 $\pm$ 2.07) & \textbf{97.74 $\pm$ 1.23} (92.62 $\pm$ 1.31) & \textbf{99.60 $\pm$ 0.17} (98.27 $\pm$ 0.20) \\
Places$365$          & \textbf{70.80 $\pm$ 2.27} (73.35 $\pm$ 3.61) & \textbf{81.74 $\pm$ 1.34} (80.52 $\pm$ 0.85) & \textbf{95.32 $\pm$ 0.17} (94.88 $\pm$ 0.31) \\
Texture            & \textbf{60.59 $\pm$ 2.65} ({67.71 $\pm$ 2.83}) & \textbf{83.46 $\pm$ 2.37} ({82.47 $\pm$ 1.77}) & \textbf{98.35 $\pm$ 0.06} ({98.20 $\pm$ 0.17}) \\
SVHN               & {44.81 $\pm$ 2.03} (\textbf{28.69 $\pm$ 1.49}) & {93.72 $\pm$ 1.29} (\textbf{95.55 $\pm$ 1.13}) & {98.76 $\pm$ 0.33} (\textbf{99.14 $\pm$ 0.20}) \\
LSUN             & \textbf{11.63 $\pm$ 1.47} (54.43 $\pm$ 2.15) & \textbf{97.85 $\pm$ 1.07} (91.96 $\pm$ 2.57) & \textbf{99.57 $\pm$ 0.17} (98.39 $\pm$ 0.25) \\
\midrule
\textbf{average}   & \textbf{40.10 $\pm$ 1.99} (55.31 $\pm$ 2.43) & \textbf{90.90 $\pm$ 1.46} (88.62 $\pm$ 1.52) & \textbf{98.32 $\pm$ 0.18} (97.77 $\pm$ 0.22) \\ \midrule[1pt]
\multicolumn{4}{c}{\cellcolor{greyL}Free Energy Scoring} \\
\midrule[1pt]
iSUN               & \textbf{32.61 $\pm$ 2.21} (45.41 $\pm$ 2.84) & \textbf{93.59 $\pm$ 1.45} (93.96 $\pm$ 1.85) & \textbf{98.74 $\pm$ 0.20} (98.23 $\pm$ 0.28) \\ 
Places$365$        & \textbf{72.64 $\pm$ 1.37} (74.99 $\pm$ 2.50) & \textbf{79.55 $\pm$ 0.87} (78.83 $\pm$ 0.90) & \textbf{94.67 $\pm$ 0.32} (94.26 $\pm$ 0.15) \\
Texture            & \textbf{67.36 $\pm$ 1.51} (67.39 $\pm$ 1.62) & \textbf{80.60 $\pm$ 1.70} (80.52 $\pm$ 1.14) & \textbf{97.00 $\pm$ 0.15} (96.90 $\pm$ 0.25) \\
SVHN               & \textbf{12.92 $\pm$ 2.20} (25.85 $\pm$ 2.47) & \textbf{97.49 $\pm$ 1.00} (95.26 $\pm$ 1.51) & \textbf{99.45 $\pm$ 0.10} (99.00 $\pm$ 0.10) \\
LSUN             & \textbf{33.53 $\pm$ 2.00} (46.68 $\pm$ 2.33) & \textbf{93.50 $\pm$ 1.50} (90.59 $\pm$ 1.70) & \textbf{98.59 $\pm$ 0.15} (97.96 $\pm$ 0.27) \\
\midrule
\textbf{average}   & \textbf{43.81 $\pm$ 1.85} (52.06 $\pm$ 2.35) & \textbf{88.94 $\pm$ 1.30} (87.83 $\pm$ 1.42) & \textbf{97.69 $\pm$ 0.18} (97.27 $\pm$ 0.21) \\ \bottomrule[1.5pt]      
\end{tabular}
}
\end{table}

\subsection{Experiments with different models}

{We demonstrate the effectiveness of our watermarking across various model architectures, including ResNet-18~\cite{he2016deep}, WRN-40-2, and ViT-B/16~\cite{DosovitskiyB0WZ21}. We conduct experiments on the ImageNet benchmark and summarized the results in Table~\ref{tab: vit}. As we can see, in both the softmax and free energy scoring cases, our watermarking can improve the detection performance across various models. However, the improvements after watermarking on the large-scale models (i.e., ViT-B/16) are not as remarkable as that of the small models (e.g., ResNet-18). It is because that the large-scale models themselves can already excel at OOD detection (better results without watermarking than that of ResNet-18 and WRN-40-2), so there may not remain a large space for their further improvements. }

\begin{table}[]
\centering
\caption{{The softmax scoring and the free energy scoring with/without watermarking on the ImageNet dataset, where we adopt different models including ResNet-18, WRN-40-2, and ViT-B/16. The notion $\downarrow$ ($\uparrow$) indicates smaller (larger) values are preferred.}} \label{tab: vit}
\scriptsize{
\begin{tabular}{c|ccc|ccc}
\toprule[1.5pt]
\multicolumn{1}{c|}{\multirow{2}{*}{}} & \multicolumn{3}{c|}{Softmax Scoring}           & \multicolumn{3}{c}{Free Energy Scoring}                   \\
\cline{2-7}
\multicolumn{1}{c|}{}                  & FPR95  $\downarrow$        & AUROC  $\uparrow$       & AUPR     $\uparrow$     & FPR95 $\downarrow$  & AUROC  $\uparrow$       & AUPR $\uparrow$         \\
\midrule[1pt]
ResNet-18                             & \textbf{41.85} (55.60) & \textbf{90.98} (86.64) & \textbf{98.22} (97.80) & \textbf{42.87} (53.26)             & \textbf{89.50} (86.42) & \textbf{97.83} (97.75) \\
WRN-40-2                              & \textbf{40.50} (54.93) & \textbf{91.22} (88.57) & \textbf{98.42} (97.69) & \textbf{43.23} (52.73)             & \textbf{89.10} (86.14) & \textbf{97.73} (97.15) \\
ViT-B/16                              & \textbf{31.63} (34.95) & \textbf{92.52} (91.31) & \textbf{86.77} (85.13) & \textbf{20.61} (21.64)             & \textbf{95.10} (94.95) & \textbf{90.87} (90.58) \\
\bottomrule[1.5pt]
\end{tabular}}
\end{table}

\subsection{Hyper-parameter Setups}
\label{app: hyper-parameter}

For the hyper-parameter setups in our experiments, we use random search to choose the proper $\sigma_1$ from the candidate parameter set $\{0.0,0.2,0.4,0.6,0.8,1.0,1.2,1.4,1.6,1.8,2.0\}$, and the proper $\rho$ from $\{0.0,0.02,0.05,0.07,0.1,0.2,0.5,0.7,1.0,2.0,5.0\}$. For softmax scoring, $\beta$ is chosen from $\{0.0,0.5,1.0,1.5,2.0,2.5,3.0,3.5,4.0,4.5,5.0\}$. For free energy scoring, $\beta$ is chosen from $\{0.0,0.02,0.04,0.06,0.08,0.1,0.2,0.4,0.6,0.8,1.0\}$, and $T_1, T_2$ are chosen from $\{0.1,0.2,0.3,0.4,0.5,0.6,0.7,0.8,0.9,1.0\}$. model performance is tested on validation OOD datasets that are separated from iSUN, Places$365$, Texture, SVHN, and LSUN datasets. 

{We adopt the random search with many trials by the following three steps. Step 1: we randomly select a hyperparameter (e.g., $\beta$) and fix the values of all other hyperparameters to be their current optimal values. Step 2: we choose the best $\beta$ from the candidate set. Step 3: do Steps 1-2 again. We repeat Steps 1 and 2 for 50 times in our experiments. Further, from Tables~\ref{tab: 18} to~\ref{tab: 29}, we list the performance of watermarking with different hyper-parameter settings for reference, where we fix the values of all other hyper-parameters (except for the considered one) to be their optimal values}. {Finally, we list the results on CIFAR benchmarks regarding the free energy scoring with different values of $T$ in Table~\ref{tab: energy t}.}

In the end, we summarize our choices of hyper-parameters, which we adopt in Section~\ref{sec: experiment} for the related experiments. On CIFAR benchmarks, our method is executed for 50 epochs. The initial learning rate $\alpha=0.01$ divided by $10$ after $25$ epochs. For the softmax scoring, we set $\sigma_1=0.4$, $\rho=1.0$, $\beta=3.5$ in CIFAR-$10$ and $\sigma_1=1.0$, $\rho=0.2$, $\beta=2.5$ in CIFAR-$100$; for the free energy scoring, we set $\sigma_1=0.6$, $\rho=0.7$, $\beta=0.1$, $T_1=0.2$, $T_2=0.7$ in CIFAR-$10$ and $\sigma_1=1.0$, $\rho=0.05$, $\beta=1.2$, $T_1=0.9$, $T_2=0.1$ in CIFAR-$100$. On the ImageNet benchmark, our method is executed for $10$ epochs and the initial learning rate $\alpha=0.01$ is divided by $10$ after $5$ epochs. We set $\rho=0.5$, $\sigma_1=0.2$, $\beta=1.5$ for the softmax scoring and $\rho=0.05$, $\sigma_1=0.4$, $\beta=0.1$  $T_1=T_2=0.5$ for the free energy scoring. Further, we fix $\sigma_2=0.001$ and $T=1$. 

We also utilize the tuning strategy with validation OOD data that are different from the test situation, where we adopt the tiny-ImageNet here for hyper-parameter tuning. The experimental results with softmax scoring are summarized from Tables~\ref{tab: 38} to~\ref{tab: 43}. As we can see, the optimal solutions chosen by tiny-ImageNet are very similar to the cases with validation sets separated from the test data, and the improvement after watermarking is remarkable as demonstrated in Table~\ref{tablast}. 

\vspace{-5pt}

\begin{table}[t]
\centering
\parbox{.30\linewidth}{
\centering
\scriptsize
\caption{Softmax scoring on CIFAR-$10$ with various $\sigma_1$.} \label{tab: 18}
\vspace{5pt}
{
\begin{tabular}{c|ccc}
\toprule[1.5pt]
             & FPR$95$     & AUROC       & AUPR      \\
\midrule[0.6pt]
2.00               & 42.11                  & 92.01                  & 98.24      \\
1.80               & 42.84                  & 91.89                  & 98.22      \\
1.60               & 41.41                  & 92.14                  & 98.25       \\
1.40               & 42.20                  & 91.82                  & 98.20      \\
1.20               & 41.98                  & 91.91                  & 98.20      \\
1.00               & 42.76                  & 91.95                  & 98.23 \\ 
0.80               & 43.38                  & 91.89                  & 98.21 \\
0.60               & 39.16                  & 92.89                  & 98.41 \\ 
\cellcolor{greyC}0.40               & \cellcolor{greyC}\textbf{38.66}                  & \cellcolor{greyC}\textbf{93.03}                  & \cellcolor{greyC}\textbf{98.45} \\
0.20               & 43.88         & 92.80         & 98.42 \\
0.00               & 48.71                  & 91.43                  & 98.11 \\ 
\bottomrule[1.5pt]      
\end{tabular}
}}~~
\parbox{.30\linewidth}{
\centering
\caption{Softmax scoring on CIFAR-10 with various $\rho$.} 
\scriptsize
\vspace{5pt}
{
\begin{tabular}{c|ccc}
\toprule[1.5pt]
            & FPR$95$     & AUROC      & AUPR      \\
\midrule[0.6pt]
5.00               & 60.02                  & 87.36                  & 97.15       \\
2.00               & 46.24                  & 91.61                  & 98.14       \\
\cellcolor{greyC}1.00               & \cellcolor{greyC}\textbf{39.12}                  & \cellcolor{greyC}\textbf{92.96}                  & \cellcolor{greyC}\textbf{98.42}       \\
0.70               & 41.07                  & 92.77                  & 98.40       \\
0.50               & 43.55                  & 92.38                  & 98.34       \\ 
0.20               & 42.02                  & 92.68                  & 98.38       \\
0.10               & 41.99                  & 92.77                  & 98.41       \\
0.07               & 42.13                  & 92.79                  & 98.42       \\
0.05               & 42.06                  & 92.84                  & 98.42       \\ 
0.02               & 44.35                  & 92.34                  & 98.32       \\
0.00               & 43.04                  & 92.44                  & 98.32       \\
\bottomrule[1.5pt]      
\end{tabular}
}
}~~
\parbox{.30\linewidth}{
\centering
\caption{Softmax scoring on CIFAR-10 with various $\beta$.} \label{tab: ablation beta full cifar10}
\scriptsize
\vspace{5pt}
{
\begin{tabular}{c|ccc}
\toprule[1.5pt]
             & FPR$95$     & AUROC       & AUPR     \\
\midrule[0.6pt]
5.00               & 40.11                  & 92.48                  & 98.34 \\
4.50               & 41.30                  & 92.39                  & 98.32 \\
4.00               & 41.21                  & 92.39                  & 98.33 \\
\cellcolor{greyC}3.50               & \cellcolor{greyC}\textbf{38.65}                  & \cellcolor{greyC}\textbf{92.55}                  & \cellcolor{greyC}\textbf{98.34}      \\
3.00               & 41.01                  & 92.55                  & 98.35 \\
2.50               & 39.66                  & 92.61                  & 98.35 \\
2.00               & 38.95                  & 92.98                  & 98.47 \\
1.50               & 43.89                  & 92.35                  & 98.33 \\
1.00               & 40.47                  & 93.08                  & 98.49 \\
0.50               & 44.51                  & 92.35                  & 98.30 \\
0.00               & 49.91                  & 91.46                  & 98.08 \\
\bottomrule[1.5pt]      
\end{tabular}
}
}
\end{table}

\begin{table}[t]
\centering
\parbox{.30\linewidth}{
\centering
\scriptsize
\caption{Energy scoring on CIFAR-$10$  with various $\sigma_1$.} \label{tab: ablation sigma full cifar10}
\vspace{5pt}
{
\begin{tabular}{c|ccc}
\toprule[1.5pt]
            & FPR$95$     & AUROC       & AUPR      \\
\midrule[0.6pt]
2.00        & 28.49       & 94.61       & 98.74     \\
1.80        & 28.61       & 94.56       & 98.71     \\
1.60        & 25.36       & 95.15       & 98.85     \\
1.40        & 28.74       & 94.85       & 98.77     \\
1.20        & 27.48       & 94.91       & 98.78     \\
1.00        & 26.47       & 95.02       & 98.83     \\
0.80        & 24.99       & 95.08       & 98.85     \\
\cellcolor{greyC}0.60        & \cellcolor{greyC}\textbf{24.50}       & \cellcolor{greyC}\textbf{95.29}       & \cellcolor{greyC}\textbf{98.93}     \\
0.40        & 26.40       & 94.84       & 98.78     \\
0.20        & 27.21       & 94.73       & 98.71     \\
0.00        & 27.97       & 94.85       & 98.75     \\
\bottomrule[1.5pt]      
\end{tabular}
}}~~
\parbox{.30\linewidth}{
\centering
\caption{Energy scoring on CIFAR-$10$  with various $\rho$.} 
\scriptsize
\vspace{5pt}
{
\begin{tabular}{c|ccc}
\toprule[1.5pt]
             & FPR$95$     & AUROC      & AUPR      \\
\midrule[0.6pt]
5.00               & 44.20                  & 91.18                  & 97.92 \\
2.00               & 29.82                  & 94.32                  & 98.69 \\
1.00               & 24.56                  & 95.10                  & 98.86 \\
\cellcolor{greyC}0.70               &  \cellcolor{greyC}\textbf{24.38}                  &  \cellcolor{greyC}\textbf{95.20}                  &  \cellcolor{greyC}\textbf{98.87} \\
0.50               & 25.96                  & 95.19                  & 98.85 \\
0.20               & 25.06                  & 95.32                  & 98.90 \\
0.10               & 29.21                  & 94.49                  & 98.69 \\
0.07               & 28.67                  & 94.78                  & 98.75 \\
0.05               & 28.72                  & 94.82                  & 98.81 \\ 
0.02               & 27.58                  & 94.93                  & 98.58 \\
0.00               & 25.38                  & 94.22                  & 98.78 \\
\bottomrule[1.5pt]      
\end{tabular}
}
}~~
\parbox{.30\linewidth}{
\centering
\caption{Energy scoring on CIFAR-$10$  with various  $\beta$.} 
\scriptsize
\vspace{5pt}
{
\begin{tabular}{c|ccc}
\toprule[1.5pt]
            & FPR$95$     & AUROC       & AUPR     \\
\midrule[0.6pt]
1.00               & 50.71                  & 91.23                  & 98.10 \\
0.80               & 42.48                  & 92.42                  & 98.30 \\
0.60               & 42.57                  & 92.78                  & 98.42 \\
0.40               & 33.74                  & 94.07                  & 98.69 \\
0.20               & 30.12                  & 94.57                  & 98.74 \\
\cellcolor{greyC}0.10               & \cellcolor{greyC}\textbf{23.68}                  & \cellcolor{greyC}\textbf{95.35}                  & \cellcolor{greyC}\textbf{98.90} \\
0.08               & 27.08                  & 94.87                  & 98.76 \\
0.06               & 25.63                  & 95.15                  & 98.84 \\
0.04               & 24.58                  & 95.06                  & 98.79 \\
0.02               & 25.47                  & 94.86                  & 98.73 \\
0.00               & 33.31                  & 92.37                  & 98.08 \\
\bottomrule[1.5pt]      
\end{tabular}
}
}
\end{table}

\begin{table}[t]
\centering
\parbox{.30\linewidth}{
\centering
\scriptsize
\caption{Softmax scoring on CIFAR-$100$ with various $\sigma_1$.} \label{tab: ablation sigma full se cifar100}
\vspace{5pt}
{
\begin{tabular}{c|ccc}
\toprule[1.5pt]
            & FPR$95$     & AUROC       & AUPR      \\
\midrule[0.6pt]
2.00               & 82.37                  & 76.45                  & 82.37 \\
1.80               & 79.72                  & 77.25                  & 94.50 \\
1.60               & 79.50                  & 77.69                  & 94.57 \\
1.40               & 79.22                  & 77.64                  & 94.58 \\
1.20               & 78.28                  & 78.45                  & 94.74 \\ 
\cellcolor{greyC}1.00               & \cellcolor{greyC}\textbf{76.57}                  & \cellcolor{greyC}\textbf{79.06}                  & \cellcolor{greyC}\textbf{94.94} \\ 
0.80               & 77.88                  & 78.84                  & 94.91 \\
0.60               & 76.84                  & 79.13                  & 94.99 \\
0.40               & 80.47                  & 76.59                  & 94.31 \\
0.20               &  81.80                 & 75.87                  & 94.14 \\
0.00               &  83.07                 & 73.81                  & 93.48 \\
\bottomrule[1.5pt]      
\end{tabular}
}}~~
\parbox{.30\linewidth}{
\centering
\caption{Softmax scoring on CIFAR-$100$ with various $\rho$.} 
\scriptsize
\vspace{5pt}
{
\begin{tabular}{c|ccc}
\toprule[1.5pt]
            & FPR$95$     & AUROC      & AUPR      \\
\midrule[0.6pt]
5.00               & 84.19                  & 72.43                  & 92.78 \\
2.00               & 76.44                  & 79.63                  & 95.08 \\
1.00               & 77.89                  & 78.58                  & 94.84 \\
0.70               & 76.39                  & 78.98                  & 94.91 \\
0.50               & 78.00                  & 78.82                  & 94.89 \\
\cellcolor{greyC}0.20               & \cellcolor{greyC}\textbf{75.43}                  & \cellcolor{greyC}\textbf{78.38}                  & \cellcolor{greyC}\textbf{94.73} \\
0.10               & 77.68                  & 78.63                  & 94.84 \\
0.07               & 78.07                  & 78.01                  & 96.60 \\
0.05               & 76.26                  & 77.77                  & 94.53 \\ 
0.02               & 77.79                  & 78.80                  & 94.81 \\
0.00               & 79.14                  & 79.14                  & 94.91 \\
\bottomrule[1.5pt]      
\end{tabular}
}
}~~
\parbox{.30\linewidth}{
\centering
\caption{Softmax scoring on CIFAR-$100$ with various  $\beta$.} \label{tab: ablation rho full se cifar100}
\scriptsize
\vspace{5pt}
{
\begin{tabular}{c|ccc}
\toprule[1.5pt]
          & FPR$95$     & AUROC       & AUPR     \\
\midrule[0.6pt]
5.00               & 78.55                  & 76.57                  & 94.13 \\
4.50               & 84.54                  & 72.69                  & 92.99 \\
4.00               & 79.27                  & 76.01                  & 93.95 \\
3.50               & 77.56                  & 76.55                  & 94.11 \\
3.00               & 79.98                  & 76.33                  & 94.12 \\
\cellcolor{greyC}2.50               & \cellcolor{greyC}\textbf{76.43}                  & \cellcolor{greyC}\textbf{77.28}                  & \cellcolor{greyC}\textbf{94.29} \\
2.00               & 79.84                  & 77.23                  & 94.38 \\
1.50               & 77.38                  & 78.85                  & 94.89 \\
1.00               & 76.91                  & 78.45                  & 94.76 \\
0.50               & 76.75                  & 79.14                  & 94.97 \\
0.00               & 83.25                  & 73.73                  & 93.45 \\
\bottomrule[1.5pt]      
\end{tabular}
}
}
\end{table}

\begin{table}[t]
\centering
\parbox{.30\linewidth}{
\centering
\scriptsize
\caption{Energy scoring on CIFAR-$100$ with various $\sigma_1$.} \label{tab: ablation sigma full cifar100}
\vspace{5pt}
{
\begin{tabular}{c|ccc}
\toprule[1.5pt]
            & FPR$95$     & AUROC       & AUPR      \\
\midrule[0.6pt]
2.00               & 77.26                  & 80.38                  & 95.32 \\
1.80               & 76.55                  & 80.56                  & 95.31 \\
1.60               & 78.09                  & 79.68                  & 95.12 \\
1.40               & 76.68                  & 80.82                  & 95.41 \\
1.20               & 75.69                  & 80.64                  & 95.34 \\ 
\cellcolor{greyC}1.00               & \cellcolor{greyC}\textbf{75.57}                  & \cellcolor{greyC}\textbf{80.80}                  & \cellcolor{greyC}\textbf{95.40} \\ 
0.80               & 75.87                  & 80.46                  & 95.25 \\
0.60               & 78.11                  & 80.12                  & 95.23 \\
0.40               & 76.18                  & 80.73                  & 95.36 \\
0.20               & 76.17                  & 80.11                  & 95.22 \\
0.00               & 75.01                  & 80.78                  & 95.39 \\
\bottomrule[1.5pt]      
\end{tabular}
}}~~
\parbox{.30\linewidth}{
\centering
\caption{Energy scoring on CIFAR-100 with various $\rho$.} 
\scriptsize
\vspace{5pt}
{
\begin{tabular}{c|ccc}
\toprule[1.5pt]
           & FPR$95$     & AUROC      & AUPR      \\
\midrule[0.6pt]
5.00               & 77.39                  & 79.91                  & 95.17 \\
2.00               & 81.39                  & 78.07                  & 94.70 \\
1.00               & 76.58                  & 80.64                  & 95.37 \\
0.70               & 78.42                       & 78.40                        & 94.76       \\
0.50               & 76.52                  & 80.54                  & 95.33 \\
0.20               & 75.32                  & 91.08                  & 95.48 \\
0.10               & 75.48                  & 80.63                  & 95.35 \\
0.07               & 75.72                  & 79.45                  & 95.03 \\
\cellcolor{greyC}0.05               & \cellcolor{greyC}\textbf{75.18}                  & \cellcolor{greyC}\textbf{79.54}                  & \cellcolor{greyC}\textbf{95.06} \\ 
0.02               & 75.83                  & 79.45                  & 95.02 \\
0.00               & 76.49                  & 79.10                  & 94.98 \\
\bottomrule[1.5pt]      
\end{tabular}
}
}~~
\parbox{.30\linewidth}{
\centering
\caption{Energy scoring on CIFAR-100 with various $\beta$.} \label{tab: 29}
\scriptsize
\vspace{5pt}
{
\begin{tabular}{c|ccc}
\toprule[1.5pt]
         & FPR$95$     & AUROC       & AUPR     \\
\midrule[0.6pt]
2.00               & 82.20                  & 78.18                  & 94.77 \\
1.80               & 79.30                  & 80.01                  & 95.20 \\
1.60               & 75.19                  & 80.53                  & 95.37 \\
1.40               & 78.39                  & 80.22                  & 95.26 \\
\cellcolor{greyC}1.20               & \cellcolor{greyC}\textbf{74.78}                  & \cellcolor{greyC}\textbf{81.51}                  & \cellcolor{greyC}\textbf{95.57} \\
1.00               & 77.68                  & 80.65                  & 95.36 \\
0.80               & 77.84                  & 90.11                  & 95.20 \\
0.60               & 75.25                  & 81.47                  & 95.60 \\
0.40               & 78.39                  & 78.67                  & 94.86 \\
0.20               & 80.46                  & 77.32                  & 94.45 \\
0.00               & 93.23                  & 73.73                  & 93.39 \\
\bottomrule[1.5pt]      
\end{tabular}
}
}
\end{table}

\begin{table}[t]
\centering
\caption{{Energy scoring on CIFAR benchmarks with various value of the hyper-parameter $T$.}} \label{tab: energy t}
\scriptsize{
\begin{tabular}{c|ccccccc|ccccccc} 
\toprule[1.5pt]
\multirow{2}{*}{$T$} & \multicolumn{7}{c|}{CIFAR-10}                                   & \multicolumn{7}{c}{CIFAR-100}      \\
\cline{2-15}
                   & 1              & 5     & 10    & 50    & 100   & 500   & 1000  & 1 & 5 & 10 & 50 & 100 & 500 & 1000 \\
\midrule[1pt]
FPR95              & \textbf{25.9} & 27.8 & 27.7 & 28.9 & 28.4 & 28.0 & 31.2 & \textbf{74.1}  & 80.4 & 77.7 & 82.3 & 80.4 & 87.0 & 89.3 \\
AUROC              & \textbf{95.0} & 94.3 & 94.0 & 93.5 & 93.5 & 93.7 & 93.4 & \textbf{81.9}  & 76.2 & 76.4 & 72.4 & 73.5 & 70.8 & 68.5 \\
AUPR               & \textbf{98.7} & 98.5 & 98.4 & 98.2 & 98.2 & 98.2 & 98.1 & \textbf{95.7}  & 94.1 & 94.1 & 92.7 & 93.0 & 91.9 & 90.9 \\
\bottomrule[1.5pt]
\end{tabular}}
\end{table}

\begin{table}[t]
\centering
\parbox{.30\linewidth}{
\centering
\scriptsize
\caption{Softmax scoring on CIFAR-$10$ with various $\sigma_1$, tiny-ImageNet is adopted as the validation set.} \label{tab: 38}
\vspace{5pt}
{
\begin{tabular}{c|ccc}
\toprule[1.5pt]
             & FPR$95$     & AUROC       & AUPR      \\
\midrule[0.6pt]
2.00               & 71.85                  & 80.50                  & 94.87      \\
1.80               & 68.50                  & 82.06                  & 95.47      \\
1.60               & 68.90                  & 83.05                  & 95.78       \\
1.40               & 66.50                  & 83.28                  & 95.91      \\
1.20               & 69.70                  & 83.39                  & 96.04      \\
1.00               & 67.30                  & 84.10                  & 96.08 \\ 
0.80               & 66.00                  & 84.75                  & 96.36 \\
0.60               & {67.30}                  & \textbf{83.92}                  & 95.97 \\ 
\cellcolor{greyC}0.40               & \cellcolor{greyC}\textbf{60.55}                  & \cellcolor{greyC}\textbf{86.81}                  & \cellcolor{greyC}\textbf{96.79}  \\
0.20               & 62.50                  & 86.21                  & 96.71  \\
0.00               & 62.05                  & 86.14                  & 96.57  \\ 
\bottomrule[1.5pt]      
\end{tabular}
}}~~
\parbox{.30\linewidth}{
\centering
\caption{Softmax scoring on CIFAR-10 with various $\rho$, tiny-ImageNet is adopted as the validation set.} 
\scriptsize
\vspace{5pt}
{
\begin{tabular}{c|ccc}
\toprule[1.5pt]
            & FPR$95$     & AUROC      & AUPR      \\
\midrule[0.6pt]
5.00               & 66.85                  & 83.89                  & 95.96       \\
2.00               & 66.15                  & 83.95                  & 95.97       \\
1.00               & 63.65                  & {84.32}                  & {96.01}       \\
0.70               & 64.50                  & 85.07                  & 96.31       \\
\cellcolor{greyC}0.50               & \cellcolor{greyC}\textbf{62.20}                  & \cellcolor{greyC}\textbf{85.72}                  & \cellcolor{greyC}\textbf{96.61}       \\ 
0.20               & 62.35                  & 85.70                  & 96.50       \\
0.10               & 63.60                  & 84.59                  & 96.04       \\
0.07               & 62.30                  & 85.61                  & 96.53       \\
0.05               & 62.65                  & 85.45                  & 96.31       \\ 
0.02               & 64.75                  & 85.54                  & 96.28       \\
0.00               & 63.80                  & 85.12                  & 96.36       \\
\bottomrule[1.5pt]      
\end{tabular}
}
}~~
\parbox{.30\linewidth}{
\centering
\caption{Softmax scoring on CIFAR-10 with various $\beta$, tiny-ImageNet is adopted as the validation set.}
\scriptsize
\vspace{5pt}
{
\begin{tabular}{c|ccc}
\toprule[1.5pt]
             & FPR$95$     & AUROC       & AUPR     \\
\midrule[0.6pt]
5.00               & 66.85                  & 83.89                  & 95.96 \\
4.50               & 66.15                  & 83.95                  & 95.97 \\
4.00               & 66.51                  & 84.32                  & 96.01 \\
3.50               & 64.50                  & 85.07                  & 96.31      \\
\cellcolor{greyC}3.00               & \cellcolor{greyC}\textbf{62.20}                  & \cellcolor{greyC}\textbf{86.72}                  & \cellcolor{greyC}\textbf{96.71} \\
2.50               & 62.35                  & 85.76                  & 96.50 \\
2.00               & 63.60                  & 84.59                  & 96.04 \\
1.50               & 64.30                  & 84.36                  & 95.94 \\
1.00               & 62.29                  & 86.00                  & 96.22 \\
0.50               & 63.05                  & 86.09                  & 96.53 \\
0.00               & 64.10                  & 86.10                 & 96.08       \\
\bottomrule[1.5pt]      
\end{tabular}
}
}
\end{table}

\begin{table}[t]
\centering
\parbox{.30\linewidth}{
\centering
\scriptsize
\caption{Softmax scoring on CIFAR-$100$ with various $\sigma_1$, tiny-ImageNet is adopted as the validation set.} 
\vspace{5pt}
{
\begin{tabular}{c|ccc}
\toprule[1.5pt]
             & FPR$95$     & AUROC       & AUPR      \\
\midrule[0.6pt]
2.00               & 83.15                  & 72.88                  & 92.82      \\
1.80               & 82.70                  & 73.34                  & 92.88      \\
1.60               & 85.60                  & 70.89                  & 92.09       \\
1.40               & 87.40                  & 69.47                  & 91.73      \\
1.20               & 87.65                  & 68.39                  & 91.42      \\
1.00               & 83.90                  & 70.46                  & 92.10 \\ 
0.80               & 83.95                  & 71.33                  & 92.17 \\
0.60               & 84.50                  & 72.12                  & 92.57 \\ 
0.40               & {83.45}                  & {72.86}                  & {92.81} \\
\cellcolor{greyC}0.20               & \cellcolor{greyC}\textbf{82.50}         & \cellcolor{greyC}\textbf{73.34}         & \cellcolor{greyC}\textbf{92.95} \\
0.00               & 83.65                       & 72.75                        & 92.20     \\ 
\bottomrule[1.5pt]      
\end{tabular}
}}~~
\parbox{.30\linewidth}{
\centering
\caption{Softmax scoring on CIFAR-100 with various $\rho$, tiny-ImageNet is adopted as the validation set.} 
\scriptsize
\vspace{5pt}
{
\begin{tabular}{c|ccc}
\toprule[1.5pt]
            & FPR$95$     & AUROC      & AUPR      \\
\midrule[0.6pt]
5.00               & 88.95                  & 64.30                  & 89.87       \\
2.00               & 84.25                  & 73.18                  & 93.03       \\
\cellcolor{greyC}1.00               & \cellcolor{greyC}\textbf{80.75}                  & \cellcolor{greyC}\textbf{74.36}                  & \cellcolor{greyC}\textbf{93.40}       \\
0.70               & 83.70                  & 72.78                  & 92.84       \\
0.50               & 83.15                  & 73.77                  & 93.18       \\ 
0.20               & 82.25                  & 73.54                  & 93.10       \\
0.10               & 81.55                  & 74.28                  & 93.37       \\
0.07               & 81.55                  & 73.75                  & 93.17       \\
0.05               & 82.35                  & 73.74                  & 93.15       \\ 
0.02               & 82.15                  & 74.12                  & 93.20       \\
0.00               & 82.70                       & 73.09                        & 93.12             \\
\bottomrule[1.5pt]      
\end{tabular}
}
}~~
\parbox{.30\linewidth}{
\centering
\caption{Softmax scoring on CIFAR-100 with various $\beta$, tiny-ImageNet is adopted as the validation set.} \label{tab: 43}
\scriptsize
\vspace{5pt}
{
\begin{tabular}{c|ccc}
\toprule[1.5pt]
             & FPR$95$     & AUROC       & AUPR     \\
\midrule[0.6pt]
5.00               & 82.85                  & 72.99                  & 92.81 \\
4.50               & 83.80                  & 73.42                  & 93.03 \\
4.00               & 81.45                  & 74.38                  & 93.33 \\
\cellcolor{greyC}3.50               & \cellcolor{greyC}\textbf{81.15}                  & \cellcolor{greyC}\textbf{74.90}                  & \cellcolor{greyC}\textbf{93.80}      \\
3.00               & 83.90                  & 73.46                  & 93.18 \\
2.50               & 81.55                  & 74.28                  & 93.20 \\
2.00               & 81.75                  & 74.75                  & 93.51 \\
1.50               & 81.20                  & 74.44                  & 93.16 \\
1.00               & 81.70                  & 74.40                  & 93.40 \\
0.50               & 82.20                  & 73.72                  & 93.34 \\
0.00               & 82.60                 & 74.81                        & 93.23       \\
\bottomrule[1.5pt]      
\end{tabular}
}
}
\end{table}


\begin{table}[t]
\centering
\caption{The softmax scoring with/without watermarking on CIFAR benchmarks. Tiny-ImageNet is adopted as the validation set for hyper-parameter tuning.} \label{tablast}
\vspace{5pt}
\scriptsize
{
\begin{tabular}{c|ccc}
\toprule[1.5pt]
                   & FPR95 $\downarrow$     & AUROC $\uparrow$       & AUPR $\uparrow$      \\
                   \cline{2-4} 
\multirow{-2}{*}{} & \multicolumn{3}{c}{w/ (w/o) watermark} \\
\midrule[0.6pt]
\multicolumn{4}{c}{\cellcolor{greyL}CIFAR-10} \\
\midrule[0.6pt]
iSUN               & \textbf{29.75} (55.00)& \textbf{95.16} (89.69)& \textbf{99.00} (97.70)\\
Places$365$        & 65.85 (\textbf{60.10})& 85.31 (\textbf{87.97})& 96.49 (\textbf{97.09})\\
Texture            & \textbf{37.05} (59.60)& \textbf{93.16} (88.43)& \textbf{98.45} (97.15)\\
SVHN               & \textbf{37.15} (46.70)& \textbf{93.99} (92.24)& \textbf{98.75} (98.34)\\
LSUN               & \textbf{28.95} (50.75)& \textbf{95.32} (91.46)& \textbf{99.04} (98.14)\\
\midrule
\textbf{average}   & \textbf{39.75} (54.43)& \textbf{92.59} (89.96)& \textbf{98.35} (97.68)\\ \midrule[1pt]
\multicolumn{4}{c}{\cellcolor{greyL}CIFAR-100} \\
\midrule[1pt]
iSUN               & \textbf{81.35} (82.30)& \textbf{75.90} (75.78)& \textbf{94.33} (94.15)\\ 
Places$365$        & \textbf{80.75} (82.90)& \textbf{74.50} (74.28)& \textbf{93.49} (93.21)\\
Texture            & \textbf{68.30} (83.55)& \textbf{77.78} (73.30)& \textbf{94.09} (92.91)\\
SVHN               & \textbf{82.60} (84.75)& \textbf{78.31} (70.64)& \textbf{95.16} (92.66)\\
LSUN               & 84.15 (\textbf{81.85})& \textbf{75.90} (74.86)& \textbf{94.35} (93.86)\\
\midrule
\textbf{average}   & \textbf{79.43} (83.07)  & \textbf{76.47} (73.77)     & \textbf{94.28} (93.35)             \\ \bottomrule[1.5pt]      
\end{tabular}
}
\end{table}

\subsection{Different Areas in the Watermark} 

\begin{figure}[t]
    \centering
    \subfigure[]{
    \centering  
    \begin{minipage}[t]{0.15\textwidth}
        \centering
	   \includegraphics[width=\linewidth]{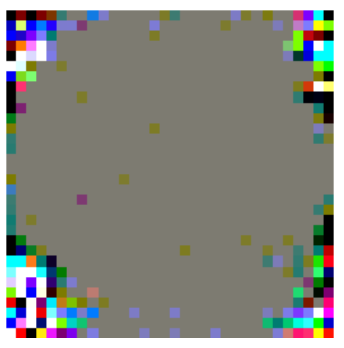}
    \centering 
    \end{minipage}} 
    ~~~~~~~~~~~~~~~~~~~~~~~~~~~~~~~~~~~~
    \subfigure[]{
    \centering  
    \begin{minipage}[t]{0.15\textwidth}
        \centering  
        \includegraphics[width=\linewidth]{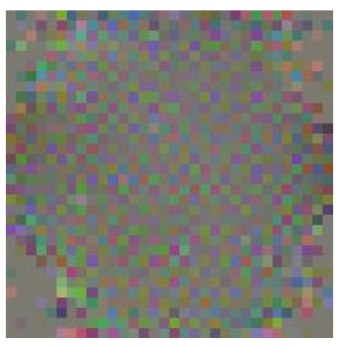}
        \centering 
    \end{minipage}}

    \caption{{Masked watermarks for the free energy scoring, with (a) $\chi_1=1$ and (b) $\chi_2=1$.}}
    \label{fig:masked}

\end{figure}

\begin{table}
    \centering
\caption{{The average performance of the free energy scoring on CIFAR-$10$ with masking. $\downarrow$ ($\uparrow$) indicates smaller (larger) values are preferred.}} \label{tab: masked}
\vspace{5pt}
\scriptsize
{
\begin{tabular}{c|ccc}
\toprule[1.5pt]
& FPR$95$ $\downarrow$     & AUROC $\uparrow$       & AUPR $\uparrow$      \\
\midrule[0.6pt]
w/ watermark & \textbf{23.69} & \textbf{95.21} & \textbf{98.83} \\
w/o watermark & 37.67 & 90.56 & 97.46 \\
\midrule[0.6pt]
\multicolumn{4}{c}{\cellcolor{greyL}$\chi_1$} \\
\midrule[0.6pt]
0.10 & 30.46 & 94.41 & 98.68 \\
1.00 & 37.72 & 93.00 & 98.52 \\
10.0 & 37.66 & 91.87 & 97.94 \\
\midrule[0.6pt]
\multicolumn{4}{c}{\cellcolor{greyL}$\chi_2$} \\
\midrule[0.6pt]
0.10 & 42.06 & 89.88 & 97.38 \\
1.00 & 51.28 & 87.54 & 96.86 \\
10.0 & 36.72 & 93.08 & 98.39 \\
\bottomrule[1.5pt]
\end{tabular}
}
\end{table}

{After watermarking, the edge area of the image is overwhelmed by the watermark's pattern, while the centre part is not much affected. However, it does not mean that the centre area is not important. Instead, under the premise of maintaining the original features, the centre area also encodes useful information in OOD detection. Table~\ref{tab: masked} is a verification of this conclusion on CIFAR-$10$ dataset with the free energy scoring, where we employ various masks in only preserving the watermark's features with their absolute values larger (smaller) than a threshold $\chi_1$ ($\chi_2$). As we can see, even if only a small portion of the watermark is masked (e.g., $\chi_1=0.10$ or $\chi_2=10.0$), there is a large drop in performance, even lower than the results without any watermarking. It indicates that both areas of the watermark contribute, and the overall watermark works as a whole for effective OOD detection. }

\end{document}